\documentclass[10pt,twocolumn,letterpaper]{article}

\usepackage[e.g.]{cvpr}
\definecolor{cvprblue}{rgb}{0.21,0.49,0.74}
\usepackage[pagebackref,breaklinks,colorlinks,allcolors=cvprblue]{hyperref}
\usepackage{makecell}
\usepackage{multirow}
\usepackage{wasysym}
\usepackage{rotating}
\usepackage[table]{xcolor}
\usepackage{array}
\def\paperID{9784} 
\def\confName{CVPR}
\def\confYear{2026}

\title{GuideFlow: Constraint-Guided Flow Matching for Planning in End-to-End Autonomous Driving}

\author{Lin Liu$^{1,2,\dagger}$, Caiyan Jia$^{1,2\ast}$, Guanyi Yu$^{3}$, Ziying Song $^{1,2\ast}$, JunQiao Li$^{3}$, Feiyang Jia$^{1,2}$,\\ Peiliang Wu$^{4}$, Xiaoshuai Hao$^{5}$, Yadan Luo$^{6}$\vspace{1ex}\\
$^1$School of Computer Science and Technology, Beijing Jiaotong University\\
$^2$Beijing Key Laboratory of Traffic Data Mining and Embodied Intelligence\\
$^3$Qcraft
$^4$Yanshan University \\
$^5$Institute of Information Engineering, Chinese Academy of Sciences\\
$^6$The University of Queensland
}

\begin{document}

\maketitle
\begin{abstract}
Driving planning is a critical component of end-to-end (E2E) autonomous driving. However, prevailing Imitative E2E Planners often suffer from multimodal trajectory mode collapse, failing to produce diverse trajectory proposals. Meanwhile, Generative E2E Planners struggle to incorporate crucial safety and physical constraints directly into the generative process, necessitating an additional optimization stage to refine their outputs. In this paper, we propose \textit{\textbf{GuideFlow}}, a novel planning framework that leverages Constrained Flow Matching. Concretely, \textit{\textbf{GuideFlow}} explicitly models the flow matching process, which inherently mitigates mode collapse and allows for flexible guidance from various conditioning signals. Our core contribution lies in directly enforcing explicit constraints within the flow matching generation process, rather than relying on implicit constraint encoding. Crucially, \textit{\textbf{GuideFlow}} unifies the training of the flow matching with the Energy-Based Model (EBM) to enhance the model's autonomous optimization capability to robustly satisfy physical constraints. Secondly, \textit{\textbf{GuideFlow}} parameterizes driving aggressiveness as a control signal during generation, enabling precise manipulation of trajectory style. Extensive evaluations on major driving benchmarks (Bench2Drive, NuScenes, NavSim and ADV-NuScenes) validate the effectiveness of \textit{\textbf{GuideFlow}}. Notably, on the NavSim test hard split (Navhard), \textit{\textbf{GuideFlow}} achieved SOTA with an EPDMS score of 43.0. The code will be released in \textcolor{blue}{https://github.com/liulin815/GuideFlow}.
\end{abstract}
\section{Introduction}
\label{sec:intro}



\begin{figure}[t]
    \centering
    \includegraphics[width=1.0\linewidth]{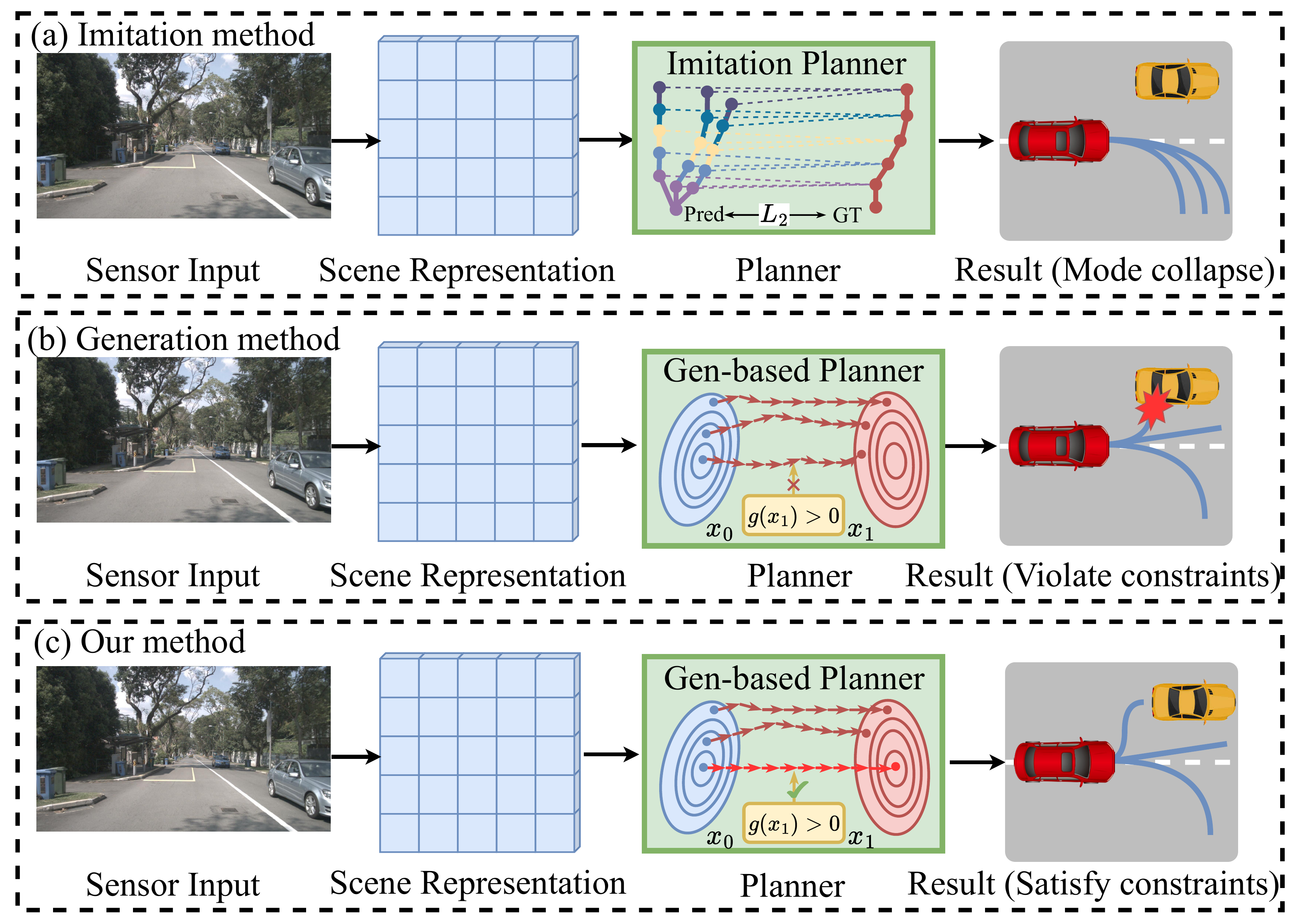}
    \caption{\textbf{Comparison of GuideFlow with prior methods.} (a) \textbf{Imitative E2E Planners}~\cite{uniad,momad,sun2024sparsedrive,jiang2023vad}, which directly imitate expert trajectories using an L2 loss, are susceptible to the inherent mode collapse problem in imitation learning. (b) \textbf{Generative E2E Planners}~\cite{liao2024diffusiondrive,goalflow}. These methods sample future trajectories directly from a learned distribution but lack explicit generation constraints, often resulting in traffic violations. (c) \textbf{GuideFlow} directly guides the generative process with explicit constraints, ensuring the sampled trajectories satisfy specific requirements.}
    \label{fig:com}
    \vspace{-1em}
\end{figure}
\let\thefootnote\relax\footnotetext{$^\dagger$ Intern of Qcraft, $^{\star}$ Corresponding author.}
In recent years, end-to-end autonomous driving (E2E-AD) \cite{uniad,LiHongyange2etpamisurvey} has emerged as a compelling alternative to traditional modular pipelines. Instead of separately optimizing perception, prediction, and planning, E2E-AD formulates the entire driving process as a single differentiable system that can be trained holistically from data. Representative frameworks such as UniAD \cite{uniad,sun2024sparsedrive,HE-Drive,jiang2023vad} exemplify this direction by coupling spatial perception \cite{bevformer,liu2022petr,bevdepth,song2024graphbev,bevfusion-pku,yang2023bevheight}, online mapping \cite{Vectormapnet,maptr,hao2024mapdistill,hao2025safemap,hao2025mapfusion,hao2025really}, motion prediction \cite{shi2022mtr,shi2024mtr++,tpnet,xu2023eqmotion}, and control decision-making \cite{bevplanner,diffusionplanner} within one coherent architecture. This joint paradigm enables cross-task reasoning and mitigates the cascading errors common in stage-wise designs. At its core, the planning module forecasts feasible, goal-directed trajectories that ultimately determine vehicle behavior.

Recent advances in E2E-AD planning have evolved from single-modal to multimodal trajectory generation to better reflect the inherent uncertainty of real-world driving \cite{li2024hydra,chen2024vadv2,sun2024sparsedrive,momad,liao2024diffusiondrive,diver}. In many scenarios, multiple plausible driving intentions often coexist, yet single-modal E2E-AD planners \cite{uniad,jiang2023vad,zheng2024genad,fusionad} produce only one deterministic path, which limits its robustness. In contrast, multimodal E2E-AD planning methods \cite{liao2024diffusiondrive,chen2024vadv2,sun2024sparsedrive,momad,FUMP,diver} instead predict multiple candidate trajectories, providing richer intent representation. However, most of these approaches are still trained under imitation learning (IL) as depicted in Fig. \ref{fig:com}. Because each driving scene provides only a single ground-truth (GT) trajectory, the learned multimodal outputs tend to collapse toward one dominant mode, resulting in highly similar predictions despite being nominally diverse. This phenomenon is commonly referred to as \textit{mode collapse}. To mitigate mode collapse, recent works~\cite{diffvla,liao2024diffusiondrive,goalflow} have explored \textit{generative modeling} for trajectory planning. Generative (Flow matching and Diffusion) approaches aim to represent the full distribution of feasible futures, where iterative sampling naturally enables diverse trajectory hypotheses. Although generative methods improve multimodal trajectory prediction, the randomness and high variance inherent in the sampling process pose a fundamental challenge to guaranteeing that generated trajectories satisfy hard safety constraints. Current approaches have rarely explored integrating explicit style and safety guidance into the generation process to ensure constraint satisfaction, posing challenges for reliable deployment.

To tackle these issues, we propose GuideFlow, a framework built upon a flow matching architecture whose generation process is explicitly supervised. GuideFlow mitigates mode collapse by starting from random samples and guiding the generation process with diverse conditioning signals. GuideFlow’s core innovation is a strategy for embedding safety constraints directly into the generative process: (1) \textit{\textbf{Constraining the Velocity Field (CVF)}}. We employ a predefined, constraint-adhering velocity field to actively correct the model's predicted velocity field, thereby steering the result to satisfy the constraints. (2) \textit{\textbf{Constraining the Flow States (CF)}}. We enforce corrections on any deviating flow paths, thereby steering flow path toward the constraint-satisfying generation endpoint. (3) \textit{\textbf{Refining the Flow by EBM (RFE)}}. By unifying flow matching architecture and EBM, we endow the model with the capacity for autonomous exploration within the data manifold, allowing it to ‘‘discover" constraint satisfying results. Our contributions are:


\begin{itemize}
    \item We propose a flow matching-based multimodal trajectory planner \textbf{GuideFlow} that effectively mitigates mode collapse. Its key innovation lies in imposing explicit hard constraints during the flow matching process and combining it with an EBM to enhance trajectory feasibility.
    \item GuideFlow employs environmental rewards as a conditioning signal, enabling switching between aggressive and conservative driving styles during inference.
    \item Extensive evaluation on autonomous driving datasets (NuScenes, ADV-NuScenes, NavSim and Bench2Drive) demonstrates its excellent performance. Notably, \textbf{GuideFlow} achieves a \textbf{SOTA} on the NavSim test hard split (Navhard) with an EPDMS score of \textbf{43.0}.
\end{itemize}

\section{Related Work}
\noindent\textbf{Imitative E2E Planners.} Imitative E2E planners~\cite{uniad,sun2024sparsedrive,HE-Drive,momad} typically regress a single expert trajectory and therefore can exhibit deterministic behaviours. UniAD~\cite{uniad} predicts multi-horizon waypoints from BEV features using pointwise regression; VAD~\cite{jiang2023vad} enhances planning with affordance cues and smoothness regularizers; TCP-Traj~\cite{e_2_e_TCP} reparameterizes trajectories with temporal control points to improve geometric stability; Drive-Adapter~\cite{jia2023driveadapter} focuses on transferring pretrained representations into the planning head;
ThinkTwice~\cite{thiktwice} adopts a coarse-to-fine refinement strategy, and Hydra-MDP~\cite{li2024hydra} introduces a high-level discrete decision layer to condition the planner. Despite architectural differences, the supervision in all these methods corresponds to a single demonstrated trajectory per scenario, optimized with Huber/L1 imitation losses.
So, these planners collapse to the dominant expert mode and cannot represent multiple plausible driving intentions in ambiguous situations.

\noindent\textbf{Generative E2E Planners.} To overcome this limitation, generative planners~\cite{liao2024diffusiondrive,goalflow,HE-Drive, diffusionplanner, diffvla} model a distribution over future trajectories rather than a single regressed path. Methods such as DiffusionDrive \cite{liao2024diffusiondrive} adopt a diffusion model but supervise only the final refinement stage, often causing sampled trajectories to converge back to a single mode. In contrast, DiffusionPlanner, Diff-VLA, and HE-Drive~\cite{diffusionplanner,diffvla,HE-Drive} explicitly supervise the denoising process and can generate diverse trajectories; however, the latent sampling process is not interpretable or constraint-aware, making it difficult to impose collision, lane, or kinematic feasibility constraints during generation.
Diff-VLA conditions each denoising step on language to guide intent, while DiffusionPlanner introduces energy-based biasing only at inference time. In contrast, GuideFlow introduces explicit supervision of the generative process combined with an energy-based model for trajectory optimization, while simultaneously enabling direct injection of hard constraints.

\begin{figure*}[t]
    \centering
    \includegraphics[width=1.01\linewidth]{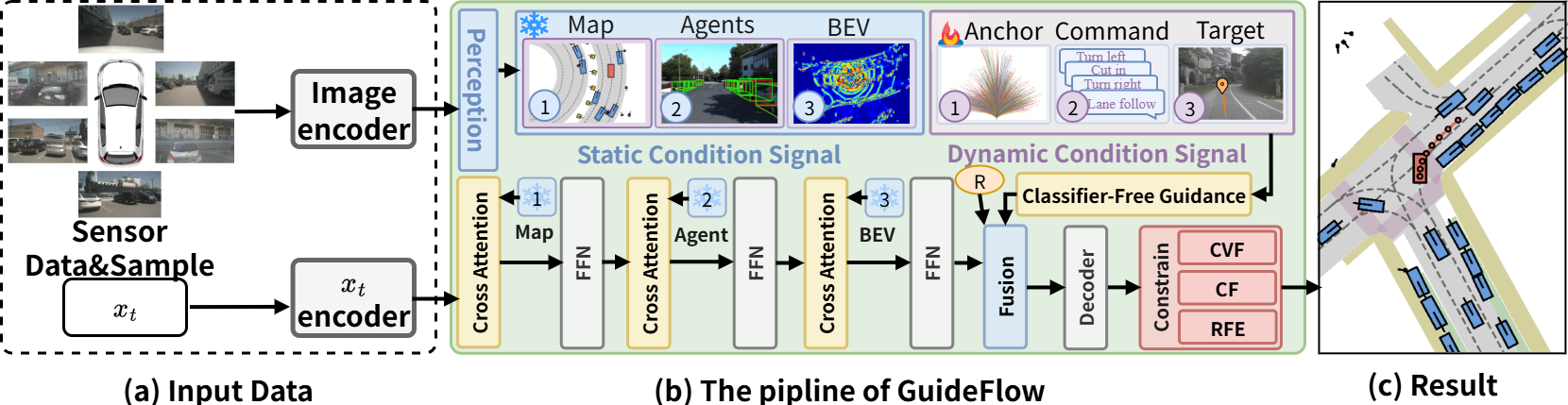}
    \caption{\textbf{The overall architecture of our proposed GuideFlow}. GuideFlow begins by encoding multi-view images into feature maps, followed by scene representation learning through the perception module. The encoded sample $x_t$ is then fused with scene representation and subjected to dynamic condition guidance for $v_{t}$ prediction. The $v_{t}$ is subsequently rectified by our proposed strategy \{CVF, CF and RFE\} ((Sec.~\ref{sec:constrain_flow})), ultimately sampling the driving trajectory. ‘‘\textcolor{orange}{R}" denotes ‘‘Reward".}
    \label{fig:main}
    \vspace{-1em}
\end{figure*}
\section{Preliminary}
\label{sec:pre}
\noindent\textbf{Flow Policy and Rectified Flow.} We start by modeling planning as flow-based trajectory generation \cite{flowmatching}, which learns a vector field to transport a simple Gaussian prior $\pi_{0}$ to the target trajectory distribution $\pi_{1}$. Let $x_t$ evolve along a probability path $\phi_t$ according to the ODE:
\begin{align}\label{eq:1}
    \frac{d x_t}{d_{t}} = v_{\theta}(x_t, t), \quad t\in[0, 1], x_0\sim \pi_0,
\end{align} 
where $v_{\theta}$ is a learnable vector field. A common instantiation is rectified flow (RF)~\cite{receified_flow}. RF constructs a linear probability path between the prior $\pi_0$ and target $\pi_1$, \textit{i.e.,} the sample $x_t = (1-t)x_{0} + tx_{1}$. Under this choice, the flow-matching learning objective is defined as :
\begin{align}\label{eq:3}
    \mathcal{L}_{\mathrm{RF}} = \mathbb{E}_{t,x_0\sim\pi_{0},x_1\sim\pi_{1}}||v_{\theta}(x_{t},t) - (x_{1} - x_{0})||^{2}.
\end{align} 
This objective efficiently learns a \textit{straight} transport toward the data manifold. At inference, trajectories are generated by numerical integration:
\begin{equation}\label{eq:flow_sampling}
    x^{(k+1)} = x^{(k)} + v_{\theta}(x^{(k)}, t_k)\Delta t, \quad x^{0}\sim\pi_0, t_k = \frac{k}{K}. 
\end{equation}
This formulation ensures fast and stable sampling, but the \textit{straight} transport path is inherently mode-seeking, often collapsing to the dominant driving pattern.

\noindent\textbf{Energy Matching}. A very recent work, Energy Matching \cite{energymatching}, introduces an energy function $E_\theta(x)$ that enables a flow model to recover multiple feasible modes. The optimality condition of the corresponding dynamic formulation is:
\begin{equation}
  \begin{aligned}
\hspace{-2ex}\frac{(x_{t+\bigtriangleup t} - x_{t})}{\bigtriangleup t} +  \bigtriangledown_{x_{t}}v_{\theta}(x_{t}) + \varepsilon(t) \bigtriangledown_{x_{t}}\log(\phi_{t}(x_t)) = 0,
\end{aligned}
  \label{eq:4}
\end{equation} 
where the energy weight schedule $\epsilon(t)$ transitions the system from pure flow transport to energy-guided manifold refinement:
\begin{equation}
    \begin{aligned}
      \varepsilon(t)=\left\{\begin{array}{ll}
      0, & 0 \leq t<\tau^{*}, \\
      \varepsilon_{\max } \frac{t-\tau^{*}}{1-\tau^{*}}, & \tau^{*} \leq t \leq 1, \\
      \varepsilon_{\max }, & t \geq 1.
      \end{array}\right.
    \end{aligned} \label{eq:varepsilon}
\end{equation}
Hear the data manifold, the transport term disappears since $x_{t+\bigtriangleup t} = x_t$, so Eq.~\eqref{eq:4} reduces to:
\begin{equation}
  \begin{aligned}
  \bigtriangledown_{x}v_{\theta}(x_{t}) + \varepsilon_{\mathrm{max}}\bigtriangledown_{x_{t}}\log(\phi_{t}(x_t)) = 0,
\end{aligned}
  \label{eq:5}
\end{equation}
This implies that the terminal distribution follows a Boltzmann form:
\begin{equation}\label{eq:6}
  \begin{aligned}
  \pi_{1}(x) \propto \exp(-\beta E_{\theta}(x)), \quad \beta = \epsilon_{\mathrm{max}}^{-1} > 0.
\end{aligned}
\end{equation}
Thus, $E_{\theta}$ shapes the manifold into multiple low-energy basins, each corresponding to a distinct feasible mode (\textit{e.g.,} yield, merge). During sampling, the discretized update becomes:
\begin{equation}\label{eq:energy_sampling}
    x^{(k+1)} = x^{(k)} + v_\theta(x^{(k)}, t_k) \Delta t - \eta(t_k) \bigtriangledown_x E_\theta(x^{(k)}),
\end{equation}
where $\eta(t)$ the discretized scheduler. In effect, the flow term efficiently transports samples towards the trajectory manifold for $0 < t < 1$, while for $t \geq \tau^*$, the energy term activates, guiding the samples into the distinct low-energy modes. This provides a principled foundation to ensure multi-modal diversity for our GuideFlow optimization.

\section{Methodology}
\label{sec:method}
To this end, we present GuideFlow as shown in Fig.~\ref{fig:main}, which acts as a flow-based trajectory generator that processes feasible and safe future motion plans. The model consists of (i) a perception-conditioned velocity field generator, (ii) classifier-free guidance that injects driving intent and style during sampling, (iii) a safety-constrained sampling procedure that operates near the data manifold via truncation and energy-based dynamics, includes: Constraining the Velocity Field (CVF), Constraining the Flow States (CF) and Refining the Flow by EBM (RFE).

\subsection{Perception-conditioned Flow Generator} \label{sec:overall_framework}
As shown in Fig.~\ref{fig:main}, Guideflow first decodes an ideal velocity field $v_{t}$ and samples feasible future trajectories $\tau$. 

\noindent\textbf{Perception to scene tokens.} Given multi-view images, we extract image features $F_{\mathrm{im}}$ and lift them into a BEV representation $F_{\mathrm{bev}}$. The perception module queries such BEV features to produce two structured token sets (1) Agent tokens $Q_{\mathrm{agent}}$ encoding interactions of dynamic agents and; (2) Map tokens $Q_{\mathrm{map}}$ embedding road and lane topology. 

\noindent\textbf{Flow state and conditioning.} We represent a trajectory at time $t$ as a flow state $x_t\in\mathbb{R}^{T\times 2}$ as in Eq.~\eqref{eq:3}, where $T$ is the prediction horizon. To condition the velocity field $v_\theta (x_t, t)$ on the scene, we map $x_t$ into a latent representation:
\begin{equation}
    h_t = \mathrm{MLP}_{\theta}(x_t) + \ell_{\theta}(t),
\end{equation}
where $\ell_{\theta}(t)$ is a sinusoidal timestep embedding. We then perform sequential cross-attention:
\begin{equation}
    \begin{aligned}
        h_t \leftarrow \operatorname{CrossAttn}_{\theta}(h_t,Q_{\mathrm{agent}}),\\
        h_t \leftarrow \operatorname{CrossAttn}_{\theta}(h_t,Q_{\mathrm{map}}).
    \end{aligned}
    \label{eq:7}
\end{equation}
Finally, we decode the velocity field $v_{\theta}(x_t, t)$ to sample future driving trajectories $\tau$:
\begin{equation}\label{eq:decode}
    v_{\theta}(x_t, t) = \mathrm{MLP}_\theta (h_t).
\end{equation}

\subsection{Classifier-free Intent and Reward Guidance}
GuideFlow incorporates high-level driving behaviors by conditioning trajectory generation on several dynamic elements that express intent and style. Specifically, we consider four possible \textit{dynamic} conditioning signals: (1) the plan anchor $C_{p}$, (2) goal point $C_{g}$, (3) driving command $C_{d}$, and (4) reward $C_{r}$ shaping the trajectory preference (as discussed in Sec.~\ref{sec:reward}). Note the driving guidance $C_p, C_g, C_d$ overlap semantically, they are not used simultaneously. 

\noindent \textbf{Implementation.} For planning anchors, we construct a trajectory vocabulary $\mathcal{V}_a$ of size $N = 256$ by applying farthest point sampling over the training set. During training, we select the plan anchors that are closest to the $gt$ trajectory as $C_p$. During sampling, GuideFlow generates $N$ trajectories by conditioning on each anchor in $\mathcal{V}_a$, enabling diverse candidate motions. Regarding the goal point $C_g$, GuideFlow derives it from the selected planning anchor. During both training and inference, GuideFlow follows the same strategy of using the planning anchor. The driving command $C_d$ is encoded as a one-hot vector for processing.

\noindent \textbf{Classifier-Free Guidance.} We adopt the classifier-free guidance   training framework \cite{classifierfreediffusionguidance}, where conditional inputs are masked $\mathcal{M}$ with a probability of $p=0.2$:
\begin{equation}
    \begin{aligned}
        h^c_t \leftarrow F_{\theta}(h_t, \mathcal{M}(C_{p} \oplus C_{g} \oplus C_{d}),\mathcal{M}(C_{r})),
    \end{aligned}
    \label{eq:8}
\end{equation}
where $F_{\theta}$ represents a cross-attention fusion module. Then the conditioned velocity field is predicted $v_{\theta}(x_t, t, c) = \mathrm{MLP}_\theta(h_t^c)$. At sampling time, we apply a guidance scale $\gamma$ to control how strongly conditions influence the motion:
\begin{equation}
    v_\theta^{\mathrm{guide}} (x_t, t, c, \gamma) = (1-\gamma)v_\theta(x_t, t) + \gamma  v_\theta(x_t, t, c).
\end{equation}

\begin{figure}
    \centering
    \includegraphics[width=1.0\linewidth]{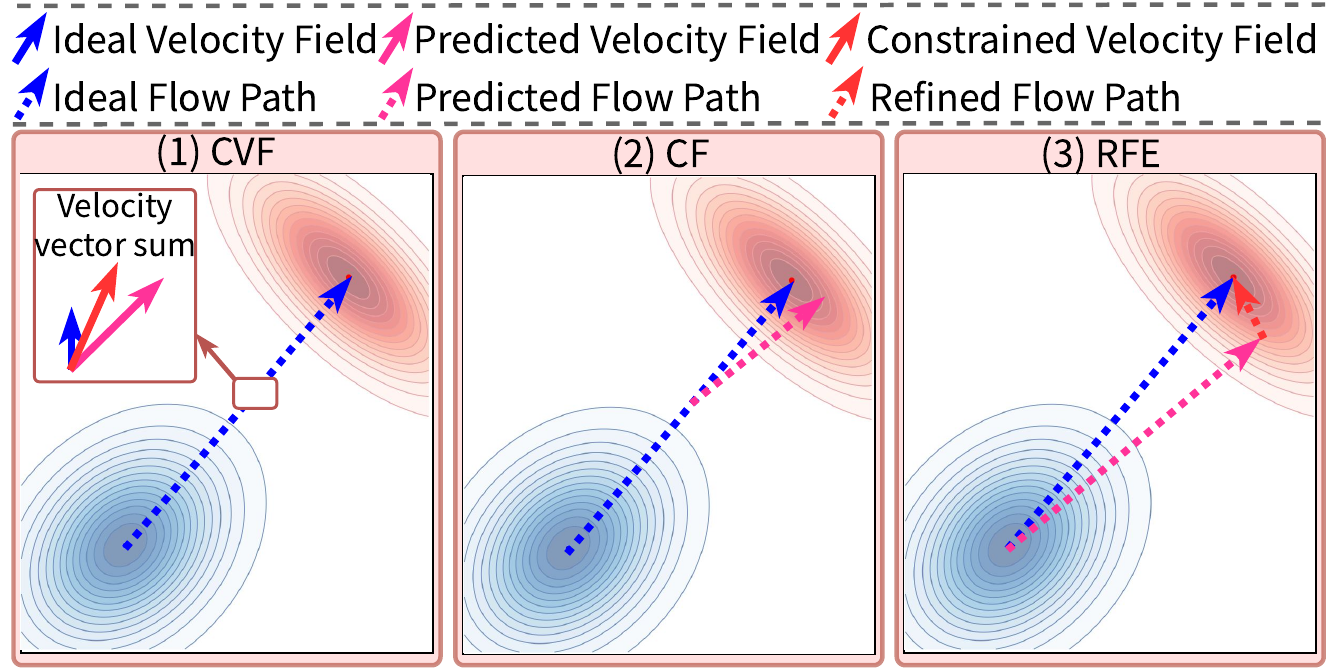}
    \caption{\textbf{Three strategies of Constrained Generation}, which include Constraining the Velocity Field (CVF), Constraining the Flow States (CF) and Refining the Flow by EBM (RFE).}
    \vspace{-1em}
    \label{fig:3}
\end{figure}

\subsection{Constrained Generation} \label{sec:constrain_flow}
While the perception and intent conditioning together enable diverse and goal-consistent motion hypotheses, they do not by themselves guarantee physical feasibility or safety. By recalling the sampling processes in Eq. \eqref{eq:flow_sampling} and Eq. \eqref{eq:energy_sampling}, it is observed that each trajectory update $x^{(k+1)}$ jointly depends on (1) the velocity field $v_\theta$, (2) the preceding flow state $x^{(k)}$ and (3)  during the refinement phase ($t>1$) the energy term $E_\theta$. This insight leads us to explore three complementary mechanisms in the following subsections, as shown in Fig.~\ref{fig:3}.

\noindent\textbf{Constraining the Velocity Field}. We first encourage the predicted motion direction to align with a constraint-satisfying reference.
Given physical or safety constraints, we manually select a feasible trajectory $x_{1}^{c}$ from the trajectory anchor set or use a pre-trained scorer (\textit{e.g.,} GTRS~\cite{GTRS}) to choose the trajectory with the highest constraint satisfaction likelihood. Its corresponding velocity field is $v_{t}^{c} = \frac{x_{1}^{c} - x_{0}}{1-0}$ between $x_{1}^{c}$ and $x_{0}$. Although potentially suboptimal, this direction ensures constraint satisfaction at the flow endpoint. To reconcile constraint compliance with motion plausibility, we synthesize a corrected velocity field:
\begin{equation}
    \begin{aligned}
        v_{t}^{*} = v_{t} - \frac{2\lambda v_{t} \cdot v_{t}^{c}}{||v_{t}^{c}||^{2}}v_{t}^{c}, \\
    \end{aligned}
    \label{eq:10}
\end{equation}
where $\lambda$ is set to 0.1 and $v_t=v_\theta(x_t, t)$ for brevity. The objective of Eq.~\eqref{eq:10} is to adjust the direction of $v_{t}$ while minimally affecting its magnitude. The proof can be found in Appendix.

\noindent\textbf{Constraining the Flow States}. While velocity-field correction aligns the overall motion direction, the flow trajectory itself may still drift away from the constraint manifold during integration. Let the continuous flow $\phi_{t}$ from $\pi_{0}$ to $\pi_{1}$, we can discretize it into a sequence ${\phi_{t}^{'}}$ according to discrete timesteps $t$:
\begin{equation}
    \begin{aligned}
        \phi_{t}^{'} = \{x^{(0)},...,x^{(k)},...,x^{(K)}\}, \quad x^{(K)}\sim\pi_1,
    \end{aligned} \label{eq:12}
\end{equation}
where $K$ is set to 100. If the generated trajectory $\tau$ fails to satisfy the constraints, it can be viewed as the $\phi_t^{'}$ deviating from the ideal flow. A straightforward correction~\cite{csp} is to manually adjust $x^{(k)}$ at each timestep to comply with constraints, but this severely disrupts the sampling process and is highly inefficient. Instead, GuideFlow adopts a \textit{truncation-like} strategy: it directly replaces discrete variables $x^{(k_{c})}$ near the target ground-truth $x_1$ with \textit{constraint-satisfying anchors} $x_{1}^{c}$ and continues sampling from it:
\begin{equation}
    \begin{aligned}
        x^{(k+1)} = x^{(k)} + v_{\theta}(x^{(k)}, t_k)\Delta t, \quad k=k_{c},...,K,
    \end{aligned}
\end{equation}
where $x^{(k_{c})} = x_{1}^{c}$ and $k_{c}$ is set to 50 in practice. In contrast to DiffusionDrive's~\cite{liao2024diffusiondrive} use of a truncation strategy in training, GuideFlow activates this mechanism only at inference, allowing the model to learn smooth conditional flows while preserving its adaptability at test time. This late-stage correction ensures the trajectories terminate in feasible regions without disrupting the learned transport dynamics. 

\noindent\textbf{Refining the Flow by EBM}. To further integrate the constraint enforcement into the generative process, we embed it directly into the energy landscape. Building upon  Eq. \eqref{eq:energy_sampling}, we interpret the flow-matching model as an energy-based model (EBM) for $t>1$, which encourages samples to converge towards low-energy and, \textit{at the same time}, constraint-satisfying regions. Consequently, we define an energy surrogate $E_{\theta}(x_{t})$ as,
\begin{equation}
    \begin{aligned}
        E_{\theta}(x_{t}) = ||\jmath(f_{t>1}(x_{t})) - \jmath(x_{t})||^{2},
    \end{aligned} \label{e_function}
\end{equation}
where $f_{t>1}(x_{t})$ denotes the sampling operator from Eq.~\eqref{eq:flow_sampling}, and $\jmath(\cdot)$ evalues constraint satisfaction (\textit{e.g.,} road complience and collision penalty) following \cite{diffusionplanner}. 
The derived $E_\theta$ assigns lower energy to feasible trajectories and higher energy to constraint violation, allowing the velocity field to implicitly learn constraint awareness during training. 



 
 Following the paradigm of EBM training, we define the training objective as:
\begin{equation}
        \mathcal{L}_{\mathrm{RFE}} = E_{\theta}(x^{(1)}) - E_{\theta}(x_{1}),
\end{equation}
where $x^{(1)}$ denotes the model's generated endpoint at $t=1$ and $x_1$ the target ground-truth. The essential role of $\mathcal{L}_{\mathrm{RFE}}$ is to increase the energy of samples that violate constraints while decreasing the energy of those that satisfy them, thereby guiding the velocity field toward regions with a higher probability of constraint satisfaction.

\subsection{Reward as Style Condition} \label{sec:reward}
To enable dynamic adjustment of trajectory aggressiveness during inference, we introduce an aggressiveness score (EP) based on NavSim, defined as the distance traveled along the lane centerline per unit time, with a value range of [0,1]. This score is computed online for each GT trajectory and incorporated as a conditional input to the model. By modulating the EP value, the aggressiveness of the generated trajectory can be directly controlled. In practice, setting EP near to 1 during inference causes the model to generate more aggressive driving behaviors.

\section{Experiments}
\begin{table*}[htbp]
\centering
\captionsetup{skip=2pt}           
\setlength{\intextsep}{2pt}
\caption{Planning results on the $\operatorname{NavSim}$\cite{dauner2024navsim} Navhard split. $^{\ast}$ denotes results reproduced with the official code repository or official checkpoint. The Scorer configuration is aligned with GTRS-Dense~\cite{GTRS}. ${\dagger}$ refers to the adjustment of the trajectory scoring strategy during inference. Further details can be found in the Appendix.}
\renewcommand\arraystretch{0.8}
\tabcolsep=0.8mm 
\setlength{\tabcolsep}{1.25mm}
\footnotesize
\begin{tabular}{l|lcccccccccccccc}
\toprule
Split & Method            & Backbone & Scorer & Stage & NC $\uparrow$ & DAC $\uparrow$ & DDC $\uparrow$ & TLC $\uparrow$ & EP $\uparrow$ & TTC $\uparrow$ & LK $\uparrow$ & HC $\uparrow$ & EC $\uparrow$ & EPDMS $\uparrow$ \\
\midrule
 \multirow{21}{*}{\rotatebox{90}{Navhard Two Stage }} & \multirow{2}{*}{$\operatorname{LTF}$~\cite{TransFuser}} &    \multirow{2}{*}{ResNet34}   &  \multirow{2}{*}{N}  &   Stage 1    &  96.2  &  79.5   &   99.1  &  99.5   &  84.1  &  95.1   &  94.2  &  97.5  &  79.1  &   \cellcolor{blue!7}    \\
      &                   &     &      &    Stage 2   &  77.7  &   70.2  &  84.2   &  98.0   &  85.1  &   75.6  &  45.4  &  95.7  &  75.9  & \cellcolor{blue!7} \multirow{-2}{*}{23.1} \\ \cmidrule{2-15}
      & \multirow{2}{*}{$\operatorname{DiffusionDrive}^{\ast}$~\cite{liao2024diffusiondrive}} &    \multirow{2}{*}{ResNet34}    & \multirow{2}{*}{N}  &   Stage 1    &  96.0  &  79.7   &   97.4  &  99.5   &  81.3  &  93.1   &  90.8  &  96.8  &  73.8  &  \cellcolor{blue!7}    \\
      &         &         &        &  Stage 2   &  82.1  &   72.2  &  88.5   &  98.7   &  85.1  &   78.8  &  49.2  &  89.3  &  71.2  & \cellcolor{blue!7} \multirow{-2}{*}{24.2} \\ \cmidrule{2-15}
      & \multirow{2}{*}{$\operatorname{GTRS-DP^{\ast}}$~\cite{GTRS}} &    \multirow{2}{*}{ResNet34}    & \multirow{2}{*}{N}  &   Stage 1    &  94.7  &  78.8   &   96.1  &  99.5   &  83.0  &  94.4   &  92.0  &  97.5  &  72.8  &  \cellcolor{blue!7}    \\
      &                   &      &     &    Stage 2   &  80.3  &   74.4  &  84.9   &  98.0   &  81.9  &   78.8  &  45.4  &  96.7  &  70.1  & \cellcolor{blue!7} \multirow{-2}{*}{23.8} \\ \cmidrule{2-15}
      & \multirow{2}{*}{\cellcolor{blue!7}} &    \multirow{2}{*}{\cellcolor{blue!7}}     &  \cellcolor{blue!7} &   \cellcolor{blue!7} Stage 1    &  \cellcolor{blue!7} 96.6  &  \cellcolor{blue!7} 80.5   &   \cellcolor{blue!7} 96.3  & \cellcolor{blue!7} 99.3   & \cellcolor{blue!7} 82.3  & \cellcolor{blue!7} 94.9   & \cellcolor{blue!7} 91.5  & \cellcolor{blue!7} 97.7  & \cellcolor{blue!7} 67.8  &   \multirow{2}{*}{\cellcolor{blue!7}}    \\
      &\cellcolor{blue!7}\multirow{-2}{*}{$\operatorname{GuideFlow}$}              &  \cellcolor{blue!7} \multirow{-2}{*}{ResNet34}    &  \cellcolor{blue!7} \multirow{-2}{*}{N}  &    \cellcolor{blue!7} Stage 2   & \cellcolor{blue!7} 87.3  &   \cellcolor{blue!7} 76.7  & \cellcolor{blue!7} 88.8   & \cellcolor{blue!7} 99.2   & \cellcolor{blue!7} 84.3  &  \cellcolor{blue!7} 85.1  & \cellcolor{blue!7} 49.7  & \cellcolor{blue!7} 93.1  & \cellcolor{blue!7} 44.5  & \cellcolor{blue!7} \multirow{-2}{*}{\textbf{27.1}} \\
      \cmidrule{2-15}
      & \multirow{2}{*}{$\operatorname{GTRS-Dense}$~\cite{GTRS}} &    \multirow{2}{*}{V2-99}   &  \multirow{2}{*}{Y}  &   Stage 1    &  98.7  &  95.8   &   99.4  &  99.3   &  72.8  &  98.7   &  95.1  &  96.9  &  40.4  &   \cellcolor{blue!7}    \\
      &                   &      &     &    Stage 2   &  91.4  &   89.2  &  94.4   &  98.8   &  69.5  &   90.1  &  54.6  &  94.1  &  49.7  & \cellcolor{blue!7} \multirow{-2}{*}{41.7} \\ \cmidrule{2-15}
      & \multirow{2}{*}{$\operatorname{DriveSuprim}$~\cite{drivesuprim}} &    \multirow{2}{*}{V2-99}   & \multirow{2}{*}{Y}   &   Stage 1    &  98.9  &  95.1   &   99.2  &  99.6   &  76.1  &  99.1   &  94.7  &  97.6  &  54.2  &  \cellcolor{blue!7}    \\
      &                   &     &      &    Stage 2   &  87.9  &   88.8  &  89.6   &  98.8   &  80.3  &   86.0  &  53.5  &  97.1  &  56.1  & \cellcolor{blue!7} \multirow{-2}{*}{42.1} \\
     \cmidrule{2-15}
      & \multirow{2}{*}{\cellcolor{blue!7}} &    \multirow{2}{*}{\cellcolor{blue!7}}     & \cellcolor{blue!7} &   \cellcolor{blue!7}Stage 1    &  \cellcolor{blue!7}98.8  &  \cellcolor{blue!7}95.5   &   \cellcolor{blue!7}99.1  &  \cellcolor{blue!7}99.5   &  \cellcolor{blue!7}76.0  &  \cellcolor{blue!7}99.1   &  \cellcolor{blue!7}94.4  &  \cellcolor{blue!7}97.5  &  \cellcolor{blue!7}52.4  &   \multirow{2}{*}{\cellcolor{blue!7}}    \\
      & \cellcolor{blue!7} \multirow{-2}{*}{\makecell{$\operatorname{GuideFlow+Scorer}$}} &  \cellcolor{blue!7}  \multirow{-2}{*}{ResNet34}   & \cellcolor{blue!7} \multirow{-2}{*}{Y}  &    \cellcolor{blue!7} Stage 2   &  \cellcolor{blue!7} 88.8  &  \cellcolor{blue!7} 89.4  &  \cellcolor{blue!7} 89.4   &  \cellcolor{blue!7} 98.8   & \cellcolor{blue!7} 80.3  &  \cellcolor{blue!7} 86.7  & \cellcolor{blue!7} 52.9  & \cellcolor{blue!7} 96.9  &  \cellcolor{blue!7} 56.9  &  \cellcolor{blue!7} \multirow{-2}{*}{\textbf{43.0}} \\ \cmidrule{2-15}
      & \multirow{2}{*}{$\operatorname{DiffVLA}$~\cite{diffvla}} &    \multirow{2}{*}{Vicuna-v1.5}   & \multirow{2}{*}{Y}   &   Stage 1    &  95.7  &  99.2   &   100  &  100   &  85.9  &  96.4   &  97.1  &  95  &  84.2  &  \cellcolor{blue!7}    \\
      &                   &     &      &    Stage 2   &  81.2  &   88.8  &  94.6   &  99.0   &  86.0  &   76.4  &  59.8  &  98.6  &  80.4  & \cellcolor{blue!7} \multirow{-2}{*}{45.0} \\ \cmidrule{2-15}
      & \multirow{2}{*}{\cellcolor{blue!7}} &    \multirow{2}{*}{\cellcolor{blue!7}}     & \cellcolor{blue!7} &   \cellcolor{blue!7}Stage 1    &  \cellcolor{blue!7}97.8  &  \cellcolor{blue!7} 97.1   &   \cellcolor{blue!7}100  &  \cellcolor{blue!7}100   &  \cellcolor{blue!7}81.4  &  \cellcolor{blue!7}98.5   &  \cellcolor{blue!7}91.4  &  \cellcolor{blue!7}92.8  &  \cellcolor{blue!7}34.2  &   \multirow{2}{*}{\cellcolor{blue!7}}    \\
      & \cellcolor{blue!7} \multirow{-2}{*}{\makecell{$\operatorname{GuideFlow+Scorer^{\dagger}}$}} &  \cellcolor{blue!7}  \multirow{-2}{*}{ResNet34}   & \cellcolor{blue!7} \multirow{-2}{*}{Y}  &    \cellcolor{blue!7} Stage 2   &  \cellcolor{blue!7} 87.3  &  \cellcolor{blue!7} 92.3  &  \cellcolor{blue!7} 98.0   &  \cellcolor{blue!7} 96.9   & \cellcolor{blue!7} 75.8  &  \cellcolor{blue!7} 85.5  & \cellcolor{blue!7} 59.3  & \cellcolor{blue!7} 95.4  &  \cellcolor{blue!7} 53.5  &  \cellcolor{blue!7} \multirow{-2}{*}{\textbf{46.7}} \\

\bottomrule
\end{tabular}
\vspace{-1.5em}
\label{tab_navsim_planning}
\end{table*}

\subsection{Experimental Setup}

\noindent\textbf{Datasets and Metrics.} For \textbf{Open Loop testing}, GuideFlow is evaluated on both the NuScenes~\cite{nuscenes} (NuS) and ADV-NuScnes~\cite{xu2025challenger} (ADV-NuS) datasets. The NuScenes dataset comprises 1,000 driving sequences. Each data sample includes six images and point clouds, providing a 360° field of view. we only utilize image data as model inputs. ADV NuScenes comprises 6,115 samples across 150 physically plausible adversarial driving scenarios, which encompasses various aggressive driving behaviors. For both NuS and ADV-NuS datasets, we replace $L_{2}$ metric with Collision Rate as the sole evaluation criterion. 

For \textbf{Closed Loop testing}, GuideFlow is evaluated on both the NavSim~\cite{dauner2024navsim} and Bench2Drive~\cite{xu2025challenger} datasets. Bench2Drive~\cite{jia2024bench2drive}, a closed-loop evaluation protocol under CARLA Leaderboard 2.0 for end-to-end autonomous driving. It provides an official training set, where we use the base set (1000 clips) for fair comparison with all the other baselines. We use the official 220 routes for evaluation. And NavSim~\cite{dauner2024navsim}, a planning benchmark derived from OpenScene, integrates multi-view camera and LiDAR data for 360° perception, with 2Hz annotations including HD maps and object bounding boxes. It employs non-reactive simulation and closed-loop evaluation for comprehensive planning assessment. For Bench2drive, we follow the Bench2Drive~\cite{jia2024bench2drive} dataset setting, measuring DS (Driving Score) and SR (Success Rate (\%)). For NavSim, we adopt NavSim's proposed Extended PMD Scores (EPDMS)~\cite{dauner2024navsim}, a weighted composite of sub-metrics.

\begin{table}[htbp]
\centering
\captionsetup{skip=2pt}           
\setlength{\intextsep}{0pt}
\caption{Planning results of E2E-AD Methods on the Bench2Drive~\cite{jia2024bench2drive} datasets. $\ast$ represents the model benefits from expert feature distillation~\cite{thiktwice}.}
\renewcommand\arraystretch{0.8}
\tabcolsep=0.8mm 
\setlength{\tabcolsep}{1.15mm}
\footnotesize
\begin{tabular}{l|c|c|c}
\toprule
\multirow{2}{*}{Method} & \multirow{2}{*}{Sensor} & \multicolumn{2}{c}{Close-loop Metrics} \\ \cmidrule{3-4}
                        &                         & Driving Score $\uparrow$      & Success Rate (\%) $\uparrow$      \\
\midrule
$\operatorname{UniAD}$~\cite{uniad}&6 Cams & \cellcolor{blue!7} 45.81  & \cellcolor{blue!7}  16.36  \\
$\operatorname{VAD}$~\cite{jiang2023vad}&6 Cams & \cellcolor{blue!7} 42.35  & \cellcolor{blue!7}15.00   \\
\midrule
$\operatorname{ThinkTwice}^{\ast}$~\cite{thiktwice}&6 Cams & \cellcolor{blue!7} 62.44  &  \cellcolor{blue!7} 31.23  \\
$\operatorname{DriveAdapter}^{\ast}$~\cite{jia2023driveadapter}&6 Cams & \cellcolor{blue!7} 64.22  & \cellcolor{blue!7} 33.08  \\
\midrule
$\operatorname{Hydra-Next}$~\cite{jia2023driveadapter}&2 Cams & \cellcolor{blue!7} 73.86  & \cellcolor{blue!7} 50.00  \\
\cellcolor{blue!7}$\operatorname{GuideFlow}$&\cellcolor{blue!7} 2 Cams &  \cellcolor{blue!7} \textbf{75.21}  &  \cellcolor{blue!7} \textbf{51.36}  \\
\bottomrule
\end{tabular}
\vspace{-1.5em}
\label{tab_bench2drive}
\end{table}

\noindent\textbf{Implementation Details.} We validated GuideFlow across four distinct benchmarks, ensuring fair comparison by aligning training protocols and baselines: For NavSim, TransFuser~\cite{TransFuser} served as the baseline. We trained on the NavTrain split for 100 epochs (LR: $2 \times 10^{-4}$). Multimodal trajectories were selected using the GTRS-Dense~\cite{GTRS} (with v2-99 backbone) scoring model. For NuScenes, implemented atop SparseDrive~\cite{sun2024sparsedrive} (700 training scenes), we followed its two-stage protocol. GuideFlow was initialized with the first-stage perception model and finetuned for 8 epochs (LR: $2 \times 10^{-4}$). Crucially, the ADV-NuScenes dataset was used only for out-of-domain evaluation and excluded from all training. For Bench2Drive, Hydra-Next~\cite{li2024hydra} was adopted as the baseline. We replaced its trajectory generation module with GuideFlow and trained the integrated model for 20 epochs (LR: $2 \times 10^{-4}$). More implementation details can be found in Appendix.

\begin{table*}[htbp]
\centering
\captionsetup{skip=2pt}           
\setlength{\intextsep}{0pt}
\caption{Planning results on the $\operatorname{NuScenes}$~\cite{nuscenes} and ADV-NuScenes~\cite{xu2025challenger} validation dataset. C.R denotes the Collision Rate. $^{\ast}$ denotes results reproduced with the official checkpoint.}
\renewcommand\arraystretch{0.8}
\tabcolsep=0.8mm 
\setlength{\tabcolsep}{2.7mm}
\footnotesize
\begin{tabular}{lcccccc cccc c}
\toprule
\multirow{2}{*}{$\operatorname{Method}$} & \multirow{2}{*}{$\operatorname{Input}$} & \multirow{2}{*}{$\operatorname{Backbone}$} & \multicolumn{4}{c}{$\operatorname{NuScenes\ C.R(\%)}\downarrow$} & \multicolumn{4}{c}{$\operatorname{ADV-NuScenes\ C.R(\%)}\downarrow$} & \multirow{2}{*}{$\operatorname{FPS}\uparrow$} \\
\cmidrule(lr){4-7} \cmidrule(lr){8-11}
& & & 1s & 2s & 3s & $\operatorname{Avg.}$ & 1s & 2s & 3s & $\operatorname{Avg.}$ \\
\midrule
$\operatorname{UniAD}$~\cite{uniad} & $\operatorname{Camera}$ & $\operatorname{ResNet101}$ & 0.62 & 0.58 & 0.63 & \cellcolor{blue!7}0.61 & 0.80 & 4.10 & 6.96 & \cellcolor{blue!7}3.95 & 1.8 $\operatorname{(A100)}$ \\
$\operatorname{VAD}$~\cite{jiang2023vad} & $\operatorname{Camera}$ & $\operatorname{ResNet50}$ & 0.03 & 0.19 & 0.43 & \cellcolor{blue!7}0.21 & 4.46 & 7.59 & 9.08 & \cellcolor{blue!7}7.05 & - \\
$\operatorname{MomAD}$~\cite{momad} & $\operatorname{Camera}$ & $\operatorname{ResNet50}$ & 0.01 & 0.05 & 0.22 & \cellcolor{blue!7}0.09 & - & - & - & \cellcolor{blue!7}- &  7.8 (RTX4090)\\
$\operatorname{DIVER~\cite{diver}}$ & $\operatorname{Camera}$ & $\operatorname{ResNet50}$ & - & - & - & \cellcolor{blue!7}- & 0.03 & 0.42 & 1.79 & \cellcolor{blue!7}0.75 & - \\
$\operatorname{DiffusionDrive~\cite{liao2024diffusiondrive}}$ & $\operatorname{Camera}$ & $\operatorname{ResNet50}$ & - & - & - & \cellcolor{blue!7}- & 0.06 & 1.29 & 3.64 & \cellcolor{blue!7}1.67 & - \\
\midrule
$\operatorname{SparseDrive^{\ast}}$~\cite{sun2024sparsedrive} & $\operatorname{Camera}$ & $\operatorname{ResNet50}$ & 0.01 & 0.05 & 0.18 & \cellcolor{blue!7}0.08 & \textbf{0.02} & 0.61 & 2.43 & \cellcolor{blue!7}1.02 & 9.0 $\operatorname{(RTX4090)}$ \\
\cellcolor{blue!7}$\operatorname{GuideFlow\ (ours)}$ & \cellcolor{blue!7}$\operatorname{Camera}$ & \cellcolor{blue!7}$\operatorname{ResNet50}$ & \cellcolor{blue!7}\textbf{0.00} & \cellcolor{blue!7}\textbf{0.02} & \cellcolor{blue!7}\textbf{0.18} & \cellcolor{blue!7}\textbf{0.07} & \cellcolor{blue!7}0.10 & \cellcolor{blue!7}\textbf{0.36} & \cellcolor{blue!7}\textbf{1.72} & \cellcolor{blue!7}\textbf{0.73} & \cellcolor{blue!7}3.6 $\operatorname{(RTX4090)}$ \\
\bottomrule
\end{tabular}
\label{tab_nuscenes_planning}
\end{table*}

\begin{table*}[htbp]
\centering
\captionsetup{skip=2pt}           
\setlength{\intextsep}{0pt}
\caption{Ablation studies of Dynamic Condition Signals in GuideFlow over NavSim~\cite{dauner2024navsim} HavHard Split, Bench2Drive~\cite{jia2024bench2drive}, NuScenes~\cite{nuscenes} and ADV-NuScenes~\cite{xu2025challenger}. ‘‘PA" denotes ‘‘Plan Anchor", ‘‘GP" denotes ‘‘Goal Point" and ‘‘CM" denotes ‘‘Driving Command".}
\renewcommand\arraystretch{0.8}
\tabcolsep=0.8mm 
\setlength{\tabcolsep}{2.4mm}
\footnotesize
\begin{tabular}{cccccccccc}
\toprule
\multicolumn{3}{c|}{Dynamic Condition} & \multicolumn{3}{c|}{NavSim Navhard (No Scorer)} & \multicolumn{2}{c|}{BenchDrive} & \multicolumn{1}{c|}{NuScenes} & ADV-NuScenes \\
\midrule
PA& GP& \multicolumn{1}{c|}{CM} & \makecell{Stage1 EPDMS}& \makecell{Stage2 EPDMS}& \multicolumn{1}{c|}{\makecell{EPDMS}} & Driving Score & \multicolumn{1}{c|}{Success Rate (\%)} & \multicolumn{1}{c|}{C. R(\%)} & C. R(\%)\\
\midrule
 & & & 56.7 & 40.0 &  23.1 &  73.86 &  50.00 &  0.08 & 1.02 \\
\cellcolor{blue!7} \checkmark & \cellcolor{blue!7} & \cellcolor{blue!7} & \cellcolor{blue!7} 58.9 & \cellcolor{blue!7} \textbf{48.0} & \cellcolor{blue!7} \textbf{29.0} & \cellcolor{blue!7} \textbf{75.21} & \cellcolor{blue!7} \textbf{51.36} & \cellcolor{blue!7} 0.08 & \cellcolor{blue!7} 0.74 \\
& \checkmark & & \textbf{59.6} & 45.7 & 28.6 &  74.54 &  50.45 &  0.08   &   0.80\\
& & \checkmark & 54.9 & 47.9 &  27.1 &  74.86 & 51.90 &  \textbf{0.07} &  \textbf{0.73}\\
\bottomrule     
\end{tabular}
\vspace{-0.5em}
\label{tab_condition_type}
\end{table*}

\subsection{Main Results}
\noindent\textbf{Closed Loop Results}. As shown in Tab.~\ref{tab_navsim_planning}, in Navhard Split, GuideFlow achieves 27.1 EPDMS, outperforming No Scorer methods (LTF~\cite{TransFuser} and GTRS-DP~\cite{GTRS}) across most metrics, demonstrating robust planning even without auxiliary scoring. When integrated with the Scorer, GuideFlow sets a new SOTA, achieving 43.0 EPDMS on Navhard split—exceeding prior best results by $+1.3$. As shown in Tab.~\ref{tab_bench2drive}, on Bench2Drive, GuideFlow achieves Driving Score of $75.04$ and Success Rate of 50.90\%, outperforming most end-to-end autonomous driving baselines. It demonstrates clear advantages over methods (ThinkTwice~\cite{thiktwice} and DriveAdapter~\cite{jia2023driveadapter}) based on expert knowledge distillation and the Hydra-Next baseline, validating the effectiveness of its generative approach in terms of closed-loop robustness and decision stability. The advancement in Bench2Drive and NavSim confirms the efficacy of incorporating constraint mechanisms into the generation process, which directly translates to improvements in key performance metrics. These consistent gains across benchmarks stem from GuideFlow’s core capability to explicitly incorporate safety constraints directly into the trajectory generation process, leading to systematic improvements in key planning and driving metrics such as EPDMS.

\begin{table*}[htbp]
\centering
\captionsetup{skip=2pt}           
\setlength{\intextsep}{0pt}
\caption{Ablation studies of different modules in GuideFlow over NavSim~\cite{dauner2024navsim} HavHard Split, Bench2Drive~\cite{jia2024bench2drive}, NuScenes~\cite{nuscenes} and ADV-NuScenes~\cite{xu2025challenger}. ‘‘EP" stands for Ego Progress subscore.  ‘‘CVF" denotes ‘‘Constraining the Velocity Field" module, ‘‘CF" denotes ‘‘Constraining the Flow State", ‘‘RFE" denotes ‘‘Refining the Flow by EBM" and ‘‘RAS" denotes ‘‘Reward as Style Condition".}
\renewcommand\arraystretch{0.8}
\tabcolsep=0.8mm 
\setlength{\tabcolsep}{1.95mm}
\footnotesize
\begin{tabular}{cccccccccccc}
\toprule
\multicolumn{4}{c|}{Modules} & \multicolumn{4}{c|}{NavSim Hard (No Scorer)} & \multicolumn{2}{c|}{Bench2Drive} & \multicolumn{1}{c|}{NuScenes} & ADV-NuScenes \\
\midrule
CVF   & CF   & RFE   & \multicolumn{1}{c|}{RAS} & EP & Stage1 EPDMS    & Stage2 EPDMS    & \multicolumn{1}{c|}{EPDMS}   & Drive Score    & \multicolumn{1}{c|}{Success Rate}   & \multicolumn{1}{c|}{C.R (\%)} & C.R (\%)     \\
\midrule
 &  &  & & \textbf{84.1} & 56.7 & 40.0 & 23.1 & 73.86 & 50.00 &  0.08 & 1.02 \\
 \checkmark &  &  & & 80.9 & 56.9 & 41.3 & 24.5 & 74.22 & 50.00 & 0.08 &  0.94 \\
 & \checkmark &  & & 81.7 & 57.1 & 44.7 & 25.1 & 74.67 & 50.45 & 0.07 & 0.77 \\
 & & \checkmark & & 81.1 & 53.3 & 45.3 &  25.5 & 74.90 & 50.45 & 0.07 &  0.83 \\
 \cellcolor{blue!7} & \cellcolor{blue!7} \checkmark & \cellcolor{blue!7} \checkmark & \cellcolor{blue!7}  & \cellcolor{blue!7}  79.6 &\cellcolor{blue!7} 54.9 & \cellcolor{blue!7}\textbf{47.9} & \cellcolor{blue!7}\textbf{27.1} &  \cellcolor{blue!7}\textbf{75.21} &  \cellcolor{blue!7}\textbf{51.36} &  \cellcolor{blue!7}\textbf{0.07} &  \cellcolor{blue!7}\textbf{0.73} \\
 & \checkmark & \checkmark & \checkmark &   82.3 & \textbf{60.1} & 43.7 & 26.3 & 74.46 & 50.45 & 0.08 & 0.80 \\
 \bottomrule
\end{tabular}
\label{tab_ablation}
\end{table*}

\begin{figure*}[ht]
    \centering
\includegraphics[width=1.0\linewidth]{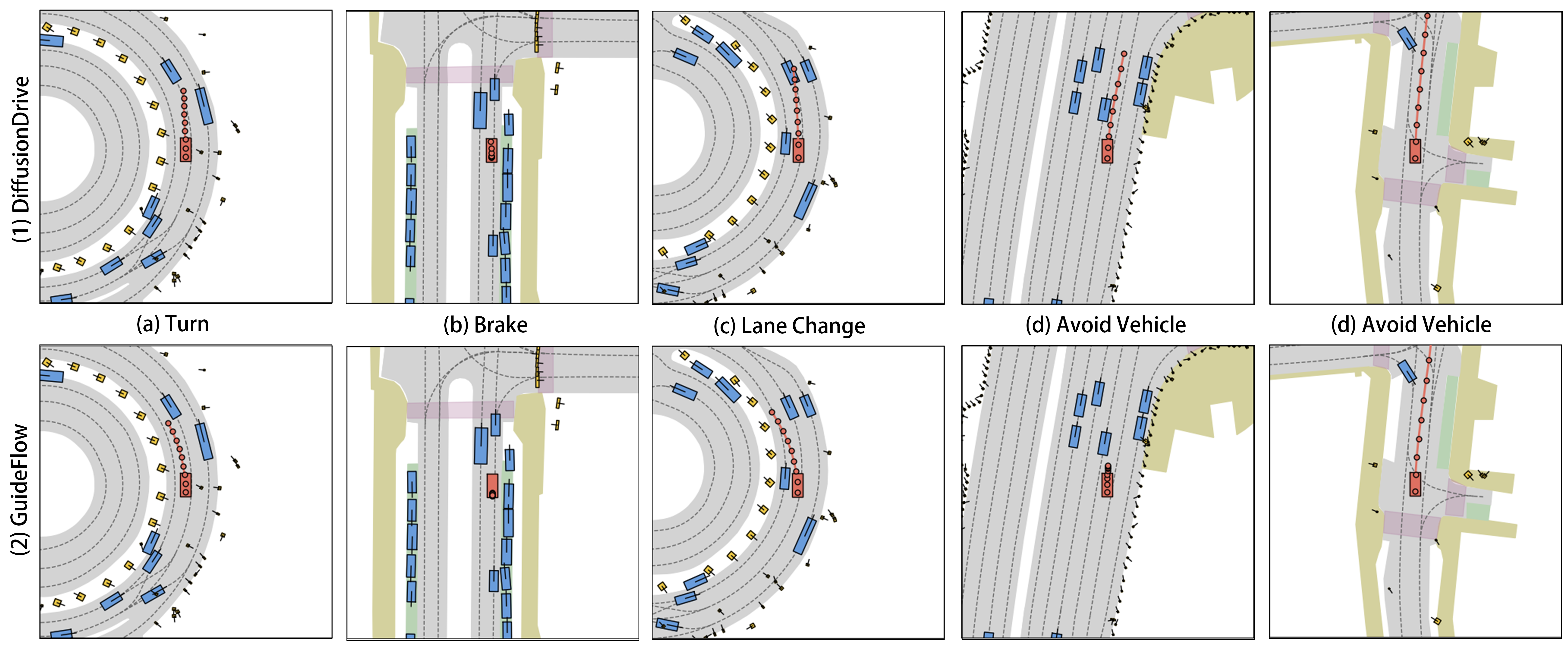}
    \caption{\textbf{Visual comparison between DiffusionDrive \cite{liao2024diffusiondrive} and our GuideFlow across multiple driving scenarios.}
GuideFlow generates trajectories that exhibit improved adherence to lane follow, smoother maneuver transitions, and stronger compliance with safety constraints such as collision avoidance and road boundary preservation.}
    \label{fig:placeholder}
\end{figure*}

\noindent\textbf{Open Loop Results}. GuideFlow is evaluated on open-loop datasets (NuScenes and ADV-NuScenes), where we use collision rate as the sole metric since the conventional L2 distance fails to properly evaluate non-imitation-based methods. As shown in Tab.~\ref{tab_nuscenes_planning}, GuideFlow achieves the lowest collision rates at all prediction horizons, demonstrating consistently safer behavior under both normal and adversarial settings. It attains an average collision rate of 0.07\% on NuScenes and 0.73\% on ADV-NuScenes, outperforming SparseDrive by 0.08\% and 1.02\%, respectively, and significantly surpassing UniAD and VAD on NuScenes. Notably, GuideFlow maintains nearly zero collisions at 1s and only 0.02\% at 2s, highlighting its short-horizon reliability. These safety gains stem directly from GuideFlow’s capacity to integrate safety constraints into the generative process, resulting in trajectories that are inherently collision-aware and robust across varying scenarios.
 
\subsection{Ablation Study}
\noindent\textbf{Effect of Different Dynamic Condition}. We conduct an ablation study on the different dynamic conditioning signals and summarize our results in Tab.~\ref{tab_condition_type}. Compared to the baseline, all model variants demonstrats performance improvements, thereby validating the efficacy of our Classifier-free Intent and Reward Guidance approach. Notably, the variant guided by the plan anchor achieves the highest performance metrics: 29.0 EPDMS and 75.21 Driving Score. This outcome surpasses the performance achieves by variants utilizing simple driving commands or goal points. This superiority stems from the plan anchor's capacity to encapsulate richer decision-making information, addressing both the intent (‘‘where to drive") and the execution (‘‘how to drive"). The results of this ablation study clearly indicate that while every individual guidance signal contributes to the overall performance enhancement within the Classifier-free Intent and Reward Guidance framework, the design of more informative and effective guidance signals remains crucial for substantially improving the model's generation capability.

\noindent\textbf{Ablations on Different Modules in Constrained Generation}. We investigated the impact of three distinct constrained generation methods on model performance, with the results summarized as shown in Tab.~\ref{tab_ablation}. Applying any of three constraint modules individually yielded performance improvements, collectively demonstrating the efficacy of the proposed Constrained Generation within GuideFlow. The CF module, in particular, delivered a more notable performance gain (+1.6 EPDMS and +0.45\% Success Rate) compared to the CVF module. This advantage is attributed to their core differences: CVF performs corrections at every generation step, which may disrupt the smoothness of the probabilistic path and degrade generation quality. In contrast, CF applies a correction only once during the generation process. This single shot intervention minimizes interference with the probabilistic path while ensuring constraint adherence, providing the model sufficient time to refine the trajectory to adjust the scene. 

Furthermore, the RFE module provides the most substantial uplift in EPDMS, particularly for out-of-domain (OOD) scenario scoring (Stage2 EPDMS). This underscores RFE's core contribution: perceiving constraint rules and guiding the model to correct the result. Because the constraint rules are fundamentally generalizable and RFE module effectively senses these rules. So GuideFlow gains the best performance in OOD scenarios. Finally, the combination of the CF and RFE modules achieves the best performance, reaching 27.1 EPDMS, 75.21 Driving Score, and a 51.36\% Success Rate. This result suggests that the methods in Constrained Generation are not antagonistic but rather complementary: CF and CVF are responsible for enforcing constraints during generation, while RFE ensures the generated output is further optimized to conform to constraint rules. More ablation studies are detailed in the Appendix.

\noindent\textbf{Ablations on Reward as Style Condition (RAS)}. In this ablation, we conduct a detailed experiment to investigate the impact of the RAS module on model performance. In the experiment, the EP reward is set to 1 to specifically encourage more aggressive trajectory. When the model incorporates the RAS module, the EP score significantly increased from 79.6 to 82.3. However, this improvement was accompanied by a 0.8 point decrease in the EPDMS score. 
This demonstrates that the indiscriminate encouragement of aggressive trajectories compromises safety constraints, resulting in a subsequent performance degradation. Nevertheless, the increased EP score confirms the feasibility of modulating trajectory aggressiveness via reward conditioning.
\begin{table}[htbp]
\centering
\captionsetup{skip=2pt}           
\setlength{\intextsep}{0pt}
\caption{The hyper-parameter $\lambda$, $k_c$ and $K$ effects on GuideFlow's performance for the NavSim Dataset.}
\renewcommand\arraystretch{0.8}
\tabcolsep=0.8mm 
\setlength{\tabcolsep}{3.6mm}
\footnotesize
\begin{tabular}{cccccc}
\toprule
$\lambda$ & \multicolumn{1}{c|}{EPDMS} & $k_c$ & \multicolumn{1}{c|}{EPDMS} & $K$ & EPDMS \\
\midrule
\cellcolor{blue!7} 0.1 & \cellcolor{blue!7} \textbf{24.5} & 10 & 24.2 & \cellcolor{blue!7} 100 & \cellcolor{blue!7} \textbf{27.1}\\
0.2 & 23.7 & 20 & 26.1 & 50 & 25.5 \\
0.3 & 18.8 & 30 & 26.3 & 25 & 23.3 \\
0.4 & 18.0 & \cellcolor{blue!7} 40 & \cellcolor{blue!7} \textbf{27.1} & 10 & 21.9\\
0.5 & 13.5 & 50 & 25.0 & - & - \\
\bottomrule
\end{tabular}
\label{tab_hy}
\vspace{-1.0em}
\end{table}

\noindent\textbf{Sensitivity of Hyper-parameters in GuideFlow}. We has ablated three key hyper-parameters, as shown in Tab.~\ref{tab_hy}: 

\noindent\textbf{Impact of $\mathbf{\lambda}$}. As $\lambda$ increases from 0.1 to 0.5, the EPMDS decreases. The performance degradation stems not from the constraint strategy itself, but from excessive interference with the predicted velocity field, which compromises the smoothness of the flow and reduces trajectory quality.

\noindent\textbf{Impact of $k_c$}. When $k_c$ increases from 10 to 50, EPMDS rises then declines. This trend suggests that CF module effectively corrects cumulative deviations, while initiating constraints too late leaves insufficient steps for the model to adapt to dynamic conditions, limiting generation quality.

\noindent\textbf{Impact of $K$}. While rectified flow's theoretically straight trajectories permit larger sampling steps, in practice, deviations from the ideal model limit the use of excessively large steps. Excessive step enlargement disrupts sampling stability, leading to erratic trajectories and performance degradation, as shown in Tab.~\ref{tab_hy}.
\section{Qualitative Results}

As shown in Fig.~\ref{fig:placeholder}, a visual comparison across diverse driving scenarios demonstrates the distinct advantages of our proposed GuideFlow method over DiffusionDrive~\cite{liao2024diffusiondrive}. Our  method successfully generates constraint-adhering trajectories, leading to a significant reduction in collision risks while maintaining strict lane discipline. Specifically, compared to DiffusionDrive, the trajectories generated by GuideFlow in Fig.~\ref{fig:placeholder} (c) and (d) clearly exhibit collision-avoidance maneuvers in response to surrounding vehicles. Furthermore, as shown in Fig.~\ref{fig:placeholder} (b), GuideFlow maintains a stationary state, preventing a potential collision with the leading vehicle. GuideFlow also demonstrates superior performance during more complex driving tasks, including lane change and turn scenarios.

\section{Conclusion}
We presents GuideFlow that leverages flow matching for planning. The core of our approach lies in its ability to incorporate diverse conditional signals, such as driving commands, goal points, and planning anchors, to guide the generation process toward context aware behaviors. Furthermore, we innovatively propose three distinct strategies to enforce explicit constraints throughout the generation process. Extensive experiments across NavSim, NuScenes, and Bench2Drive confirm GuideFlow's effectiveness. GuideFlow demonstrates superior robustness, particularly in challenging out-of-domain scenarios. While GuideFlow performs excellently, accelerated sampling can compromise its performance. Future work will integrate reflow and meanflow to enhance the model's sampling speed.

\small
\bibliographystyle{ieeenat_fullname}
\bibliography{main}
\end{document}


\clearpage
\setcounter{page}{1}
\maketitlesupplementary

\appendix

\section{Appendix}
\noindent This supplementary material provides additional descriptions of the proposed GuideFlow framework, including the following supplementary material:
\begin{itemize}
    \item \textbf{\cref{sec:Contribution}:} Summary of Contributions.
    \item \textbf{\cref{sec:proof}:} Proofs within the GuideFlow.
    \item \textbf{\cref{sec:dataset_and_metric}:} The Details of Metrics.  
    \item \textbf{\cref{sec:Implementation_Details}:}  Implementation Details.
    \item \textbf{\cref{sec:Results}:} More Ablation Studies.
    \item \textbf{\cref{sec:Qualitative}:} More Visualizations of Planning Results.
\end{itemize}

\subsection{Summary of Contributions}
\label{sec:Contribution}
Our contributions are summarized below.

\noindent 1) \textbf{GuideFlow Framework.} We propose GuideFlow, an innovative framework based on constrained flow matching. This framework explicitly models the flow matching process and incorporates diverse conditional signals to guide trajectory generation, thereby effectively mitigating the "mode collapse" issue. By explicitly embedding safety constraints into the generation process, GuideFlow ensures strict compliance of output trajectories with safety requirements. This approach significantly enhances the stability and safety of motion planning in end-to-end autonomous driving systems.

\noindent 2) \textcolor{blue}{\textbf{New$^*$}}  \textbf{CVF, CF and RFE Modules. }We propose the Constraining the Velocity Field (CVF) module, which employs a predefined, constraint-adhering velocity field to actively correct the model's predicted velocity field, thereby steering the result to satisfy the constraints. The proposed Constraining the Flow States (CF) module enforce corrections on any deviating flow paths, thereby steering flow path toward the constraint-satisfying generation endpoint. Furthermore, we propose the Refining the Flow by EBM (RFE) module. By unifying flow matching architecture and EBM, we endow the model with the capacity for autonomous exploration within the data manifold, allowing it to ‘‘discover" constraint-satisfying results.

\noindent 3) \textcolor{blue}{\textbf{New$^*$}} \textbf{Reward as Style Condition Module. } We propose the Reward as Style Condition Module. This module encodes the aggressiveness score, which evaluates trajectories in driving scenarios, into a conditional control signal. This signal enables GuideFlow to dynamically adjust the aggressiveness level of trajectories during the generation process, thereby producing driving trajectories with a broader range of behavioral styles. The details for calculating the aggressiveness score will be presented in Sec~\ref{sec:ep_r}.

\subsection{Proofs within the GuideFlow}
\label{sec:proof}
In this section, we prove the theoretical validity of the proposed RFE module.
\subsubsection{Proofs for RFE Module}

\noindent The starting point of RFE module is the JKO scheme~\cite{jko}. The JKO scheme describes the discrete-time evolution of a probability distribution $\rho_{t}$ along energy-minimizing trajectories in the Wasserstein space,
\begin{equation}
    \begin{aligned}
        \rho_{t+\Delta t} &=\arg\min_{\rho}\frac{W_{2}^{2}\left(\rho,\rho_{t}\right)}{2\Delta t} +\int V_{\theta}(x)\mathrm{d}\rho(x) \\ &+  \varepsilon(t)\int\rho(x)\log\rho(x)\mathrm{d}x.
    \end{aligned}
\end{equation}
Here, $\theta$ denotes the learnable parameters of the scalar potential $V_{\theta}(x)$, and $\varepsilon(t)$ is a temperature-like parameter tuning the entropic term. The transport cost is given by the Wasserstein distance:
\begin{equation}
\begin{aligned}
W_2^2(\rho,\rho_t)=\min_{\gamma\in\Gamma(\rho,\rho_t)}\int_{\mathbb{R}^d\times\mathbb{R}^d}\left\|x-x_t\right\|^2\mathrm{d}\gamma(x,x_t),
\end{aligned}
\end{equation}

\noindent where $\Gamma(\rho,\rho_t)$ is the set of couplings between $\rho$ and $\rho_t$, i.e., the set of probability distributions on $\mathbb{R}^d\times\mathbb{R}^d$ with marginals $\rho$ and $\rho_t$. Here, $d$ is the dimensionality of the data. Following Energy Matching~\cite{energymatching}, the $\epsilon(t)$ is designed as a time-dependent linear schedule.
\begin{equation}
    \begin{aligned}
      \varepsilon(t)=\left\{\begin{array}{ll}
      0, & 0 \leq t<\tau^{*}, \\
      \varepsilon_{\max } \frac{t-\tau^{*}}{1-\tau^{*}}, & \tau^{*} \leq t \leq 1, \\
      \varepsilon_{\max }, & t \geq 1.
      \end{array}\right.
    \end{aligned} \label{eq:varepsilon}
\end{equation}
Then, we follow the approach in \cite{terpin2024learning,energymatching} and study at each time $t$ via its first-order optimality conditions:
\begin{equation}
  \begin{aligned}
\hspace{-2ex}\frac{(x_{t+\bigtriangleup t} - x_{t})}{\bigtriangleup t} +  \bigtriangledown_{x_{t}}v_{\theta}(x_{t}) + \varepsilon(t) \bigtriangledown_{x_{t}}\log(\phi_{t}(x_t)) = 0,
\end{aligned}
  \label{eq:9}
\end{equation} 
Hear the target data manifold, the transport term disappears since $x_{t+\bigtriangleup t} = x_t$, so Eq.~\eqref{eq:9} reduces to:
\begin{equation}
  \begin{aligned}
  \bigtriangledown_{x}v_{\theta}(x_{t}) + \varepsilon_{\mathrm{max}}\bigtriangledown_{x_{t}}\log(\phi_{t}(x_t)) = 0,
\end{aligned}
  \label{eq:5}
\end{equation}
This implies that the terminal distribution follows a Boltzmann form:
\begin{equation}\label{eq:6}
  \begin{aligned}
  \pi_{1}(x) \propto \exp(-\beta E_{\theta}(x)), \quad \beta = \epsilon_{\mathrm{max}}^{-1} > 0.
\end{aligned}
\end{equation}
Thus, $E_{\theta}$ shapes the manifold into multiple low-energy basins, each corresponding to a distinct feasible mode (\textit{e.g.,} yield, merge). During sampling, the discretized update becomes:
\begin{equation}\label{eq:energy_sampling}
    x^{(k+1)} = x^{(k)} + v_\theta(x^{(k)}, t_k) \Delta t - \eta(t_k) \bigtriangledown_x E_\theta(x^{(k)}),
\end{equation}
where $\eta(t)$ the discretized scheduler. In effect, the flow term efficiently transports samples towards the trajectory manifold for $0 < t < 1$, while for $t \geq \tau^*$, the energy term activates, guiding the samples into the distinct low-energy modes. This provides a principled foundation to ensure multi-modal diversity for our GuideFlow optimization.


\subsection{The Details of Metrics}
\label{sec:dataset_and_metric}
\noindent For \textbf{NuScenes and ADV-NuScenes} datasets, we adopt the Collision Rate as the primary performance metric, while excluding the $L_{2}$ Error. This selection is justified by the observation that multiple feasible driving trajectories often exist in real-world driving scenarios. Therefore, simply evaluating the similarity between the predicted trajectory and the expert trajectory is insufficient for accurately assessing the model's multimodal trajectory generation capability. Specifically, the Collision Rate is computed over a 3s prediction horizon. The trajectory is sampled at a time interval of 0.5s and the average value is reported as the final result.

\noindent For \textbf{Bench2Drive}~\cite{jia2024bench2drive} dataset, we adopt the Success Rate (SR) and Driving Score (DS) as the primary performance metrics. The Success Rate measures the proportion of successfully completed routes within the allotted time and without traffic violations. A route is deemed successful if the ego vehicle reaches its destination without any rule infractions. The success rate is calculated as the ratio of successful routes to the total number of routes. This metric follows CARLA~\cite{CARLA} official metric as reference. It considers both route completion and penalty for infractions. Specifically, it averages the route completion percentages and penalizes infractions based on their severity as shown in . The driving score is normalized by the total number of routes from same type or group as well.

\begin{equation}
    \begin{aligned}
        \text{DS}=\frac{1}{n_{\mathrm{total}}}\sum_{i=1}^{n_{\mathrm{total}}}\mathrm{RC}_{i}*p_{i},
    \end{aligned}
\end{equation}
where $n_{\mathrm{total}}$ denotes the number of successful routes and total samples respectively; $\mathrm{RC}_{i}$ representats the percentage of route distance completed for the $i$-th route; $p_{i}$ means the infraction penalty on the $i$-th route.
 
\noindent For \textbf{NavSim}~\cite{dauner2024navsim} dataset, we employ the proposed extended PDM score (EPDMS)~\cite{dauner2024navsim}, which is a weighted combination of several sub-scores: No at-fault Collisions (NC), Drivable Area Compliance (DAC), Driving Direction Compliance (DDC), Traffic Light Compliance (TLC), Ego Progress (EP), Time to Collision (TTC) within bound, Lane Keeping (LK), History Comfort (HC) and Extended Comfort (EC).
\subsection{Implementation Details}
\label{sec:Implementation_Details}
\subsubsection{Experimental Setup}
For \textbf{NuScenes and ADV-NuScenes}, we load the first-stage pre-trained model of SparseDrive. This model trains the sparse perception module from scratch, which encompasses $3\text{D}$ object detection, multi-object tracking, and online mapping, to learn sparse scene representations. Then, we train the $3\text{D}$ object detection, online mapping, motion, and planning modules without freezing the weights of the sparse perception module of GuideFlow. For GuideFlow, we use ResNet50~\cite{resnet} as backbone network and the input image size is 256 $\times$ 704. We trained GuideFlow for $8$ epochs using a learning rate of $2 \times 10^{-4}$ and a total batch size of $48$, using 8 NVIDIA 4090 GPUs. The loss function for the supervised process during the training is defined as follows,
\begin{equation}
    \begin{aligned}
       \mathbf{L} = \mathbf{L_D} + \mathbf{L_M} + \mathbf{L_{MP}} + \mathbf{L_{RF}} + \mathbf{L_{RFE}},
    \end{aligned}
\end{equation}
Where, $\mathbf{L_{RF}}$ denotes the loss function for the Flow Matching process, and $\mathbf{L_{EM}}$ represents the loss generated during the RFE Module process. During the inference phase, GuideFlow generates $18$ trajectory proposals by performing $K=100$ sampling steps. These proposals are then selected using the scorer embedded within SparseDrive.

\noindent For \textbf{Bench2Drive}, GuideFlow is built upon the Hydra-Next baseline architecture and employs ResNet-50 as the image backbone to extract front-view and back-view image features. We train GuideFlow on the training data for 20 epochs with a
total batch size of 256 and a learning rate of $2 \times 10^{-4}$, using 8 NVIDIA V100 GPUs. And the loss function for the supervised process during the training is defined as follows,

\begin{equation}
    \begin{aligned}
       \mathbf{L} = \mathbf{L_{HY}} + \mathbf{L_{RF}} + \mathbf{L_{RFE}}.
    \end{aligned}
\end{equation}
GuideFlow replaces the original trajectory generation module of Hydra-Next. Consequently, the loss term $\mathbf{L_{HY}}$ is defined as the summation of the losses from all other modules of the original Hydra-Next architecture, excluding the trajectory generation component. During the inference phase, GuideFlow generates $10$ trajectory proposals by performing $K=100$ sampling steps.

\noindent For \textbf{NavSim}, the TransFuser serves as our baseline. We adopt ResNet-34 as the image backbone to extract visual features from the front-view, left-front view, and right-front view camera inputs. GuideFlow was trained for $100$ epochs using a total batch size of $64$ and a learning rate of $2 \times 10^{-4}$, using 8 NVIDIA V100 GPUs. And the loss function for the supervised process during the training is defined as follows,
\begin{equation}
    \begin{aligned}
       \mathbf{L} = \mathbf{L_D} + \mathbf{L_M} + \mathbf{L_{RF}} + \mathbf{L_{RFE}}.
    \end{aligned}
\end{equation}
During the inference phase, GuideFlow generates $100$ trajectory proposals by performing $K=100$ sampling steps. GTRS-Dense~\cite{GTRS} (with v2-99 backbone) serves as a scorer for trajectory selection. Furthermore, we dynamically adjusted the model's scoring rule during inference. More details regarding this adjustment are provided in Sec.~\ref{sec:opt}.
\subsubsection{Constraint Satisfaction Evaluation Function $\jmath(\cdot)$}\label{sec:cse}
In this paper, we focus on two primary constraints: first, collision avoidance with other agents in the environment, and second, guaranteeing the ego vehicle stays within drivable area. We will detail how to calculate the constraint satisfaction for trajectories.

\noindent \textbf{Collision Avoidance.}
Given a predicted trajectory $\tau \in \mathbb{R}^{T \times 2}$, we calculate the signed distance $d_{t}$ between the ego-vehicle and surrounding agents at each timestep $t$. A positive value $d_{t}^{+}$ indicates no collision (i.e., safe separation), while a negative value $d_{t}^{-}$ signifies that a collision has occurred. The satisfaction degree of the collision constraint for trajectory $\tau$ is calculated as follows:
\begin{equation}
    \begin{aligned}
        \jmath_{c}(\tau) = & \frac{\sum_{t}^{T}1_{d_{t}^{+}>0}\cdot f\left(\omega_{\mathrm{c}}\cdot\max\left(1-\frac{d_{t}^{+}}{r},0\right)\right)}{\omega_{\mathrm{c}}(\sum_{t}^{T}1_{d_{t}^{+}>0}+\mathrm{eps})} \\
        & \frac{\sum_{t}^{T}1_{d_{t}^{-}<0}\cdot f\left(\omega_{\mathrm{c}}\cdot\max\left(1-\frac{d_{t}^{-}}{r},0\right)\right)}{\omega_{\mathrm{c}}(\sum_{t}^{T}1_{d_{t}^{-}<0}+\mathrm{eps})},
   \end{aligned}
\end{equation}
where $f(\cdot) = e^{x} - x$. $r$ denotes the collision-sensitive distance and eps is added to ensure numerical stability.

\noindent \textbf{Stay within Drivable Area.}
Given the road segmentation map, we convert the drivable area and other regions into a binary mask map. This mask is used to calculate the Signed Distance Field (SDF), where the drivable area is assigned negative values and regions outside the drivable area are assigned high positive values. We project the trajectory points onto the SDF and obtain the corresponding values $d_{t}$ via nearest neighbor interpolation. Finally, the satisfaction degree of the drivable area constraint for the trajectory $\tau$ is calculated as follows:
\begin{equation}
    \begin{aligned}
\jmath_{d}(\tau) = \sum_{t}^{T}g\left(\omega_{\mathrm{d}}\cdot d_{t}\right),
   \end{aligned}
\end{equation}
where $g(\cdot) = e^{x} - x$.
\subsubsection{Optimization of the Scoring Strategy for NavSim}\label{sec:opt}
During inference on the NavSim dataset, we observed that the GTRS-Dense trajectory scorer exhibited a strong bias towards selecting trajectories from the pre-clustered anchor trajectory table, rather than the real-time generated trajectories. This behavior is attributed to the fact that GTRS-Dense was trained solely on anchor scores (due to the difficulty of obtaining accurate ground-truth scores for online generated trajectories during training). Consequently, the model was only exposed to sub-optimal anchor trajectories rather than the optimal real-time solutions. To mitigate this bias, we enforce a higher selection rate for the generated trajectories during inference by manually boosting their scores by an additive factor of $0.1$.
\subsubsection{The Computation Method for EP Reward} \label{sec:ep_r}
To quantify the aggressiveness score of trajectories, GuideFlow introduces the EP Reward. This reward is designed to evaluate the ego vehicle's progress in advancing along the lane centerline within a specified time frame. Specifically, for each trajectory, we first determine the projection positions of its start point $P_{\text{s}}$ and end point $P_{\text{e}}$ onto the reference centerline. The raw progress value is defined as the difference between the arc length of the end point projection and that of the start point projection:
\begin{equation}
    \begin{aligned}
EP = \mathbf{MAX}(0,\mathrm{proj}(P_{\text{e}})-\mathrm{proj}(P_{\text{s}})),
   \end{aligned}
\end{equation}
where the projection function $\mathrm{proj}(\cdot)$ maps points to their closest points on the centerline. $\mathrm{proj}(\cdot)$ is implemented using the \textit{shapely} Python library.
\subsection{More Ablation Studies}
\label{sec:Results}
\subsubsection{Combinations of CVF, CF, and RFE Modules}
In this section, we present additional ablation studies about CVF, CF and RFE modules, as shown in Tab.~\ref{tab_ablation_apd}. To systematically validate the effectiveness of the proposed approach, we conducted a comprehensive evaluation of different combinations of the CVF, CF, and RFE modules within the GuideFlow framework. Experimental results show that while the combination of CVF and CF improves performance compared to using CF alone, it still falls short of the performance achieved by CVF individually. This indicates that both CF and CVF perform corrections during the flow matching process, and simply combining them leads to "over-correction" issues, resulting in irregular probability flows and consequently suboptimal performance. The combination of RFE and CVF modules further confirms this finding. Since the RFE module performs optimization after the flow matching process, its integration with CVF effectively avoids multiple correction conflicts during the matching phase, thus achieving better performance than using either module alone.
\begin{table}[htbp]
\centering
\captionsetup{skip=2pt}           
\setlength{\intextsep}{0pt}
\caption{Ablation studies of different modules in GuideFlow over NavSim~\cite{dauner2024navsim} HavHard Split. ‘‘EP" stands for Ego Progress subscore.  ‘‘CVF" denotes ‘‘Constraining the Velocity Field" module, ‘‘CF" denotes ‘‘Constraining the Flow State", ‘‘RFE" denotes ‘‘Refining the Flow by EBM" and ‘‘RAS" denotes ‘‘Reward as Style Condition".}
\renewcommand\arraystretch{0.8}
\tabcolsep=0.8mm 
\setlength{\tabcolsep}{0.9mm}
\footnotesize
\begin{tabular}{cccccccc}
\toprule
\multicolumn{4}{c|}{Modules} & \multicolumn{4}{c}{NavSim Hard (No Scorer)} \\
\midrule
CVF   & CF   & RFE   & \multicolumn{1}{c|}{RAS} & EP & Stage1 EPDMS    & Stage2 EPDMS    & \multicolumn{1}{c}{EPDMS} \\
\midrule
 &  &  & & \textbf{84.1} & 56.7 & 40.0 & 23.1 \\
 \checkmark &  &  & & 80.9 & 56.9 & 41.3 & 24.5 \\
 & \checkmark &  & & 81.7 & 57.1 & 44.7 & 25.1 \\
 \checkmark & \checkmark &  & & 82.8 & 56.2 & 44.8 & 24.9 \\
 & & \checkmark & & 81.1 & 53.3 & 45.3 &  25.5 \\
 & \checkmark & \checkmark & & 79.6 & 54.9 & \textbf{47.9} & \textbf{27.1} \\
 \checkmark & & \checkmark & & 80.0 & 56.5 & 45.4 &  26.2 \\
 & \checkmark & \checkmark & \checkmark &   82.3 & \textbf{60.1} & 43.7 & 26.3 \\
 \checkmark & \checkmark & \checkmark & \checkmark &   82.5 & 60.0 & 43.7 & 26.1 \\
 \bottomrule
\end{tabular}
\label{tab_ablation_apd}
\end{table}

\begin{table}[]
\centering
\captionsetup{skip=2pt}           
\setlength{\intextsep}{0pt}
\caption{Ablation studies on the effects of ``EP Reward" parameter configuration within the RAS Module on model performance over NavSim HavHard Split.}
\renewcommand\arraystretch{0.8}
\tabcolsep=0.8mm 
\setlength{\tabcolsep}{1.15mm}
\footnotesize
\begin{tabular}{lccccclccccc}
\toprule
Reward & 0.1 & 0.2 & 0.3 & 0.4 & 0.5 & 0.6 & 0.7 & 0.8 & 0.9 & 1.0 \\
\midrule
EP     &  79.4   &   79.1  &  79.8   &   80.3  &  80.3 & 80.6 & 81.3 & 81.7 & 81.6 & 82.3 \\
EPDMS  &   25.6  &   25.4  &   25.5  &   26.0  &  26.2 & 26.3 & 26.1 & 26.8 & 26.5 & 26.3 \\
\bottomrule
\end{tabular} \label{tab_ep}
\end{table}

\subsubsection{Effect of the RAS module guidance strength}
To further investigate the effectiveness of the RAS module, we systematically examined its impact on the model by configuring different ``EP Reward" values during inference. As shown in Tab.~\ref{tab_ep}, as the ``EP Reward" increases from 0.1 to 1, the model's EP score progressively rises and eventually peaks, demonstrating the RAS module's capability to effectively guide trajectory style variations through reward signaling. 

Furthermore, we observe a notable phenomenon: EPDMS reaches its optimum at ``EP Reward" of 0.8, then decreases with further reward increase. This indicates that indiscriminately encouraging aggressive behavior is not suitable for all driving scenarios (e.g., congested road conditions). The optimal reward setting should be adaptively determined by the model rather than manually specified. Automating the configuration of reward will be a direction for our future work. Despite this observation, the ablation study robustly validate the core innovation of the RAS module: effectively guiding trajectory evolution toward desired directions through dynamic reward adjustment.

\begin{table}[]
\centering
\captionsetup{skip=2pt}           
\setlength{\intextsep}{0pt}
\caption{The hyper-parameter $\lambda$, $k_c$ and $K$ effects on GuideFlow's performance for the NuScenes and ADV-NuScenes Dataset. ``C.R." denotes ``Collison Rate".}
\renewcommand\arraystretch{0.8}
\tabcolsep=0.8mm 
\setlength{\tabcolsep}{1.3mm}
\footnotesize
\begin{tabular}{ccccccccc}
\toprule
\multirow{2}{*}{$\lambda$} & NuS                    & ADV-NuS                    & \multirow{2}{*}{$k_c$} & NuS                  & ADV-NuS                    & \multirow{2}{*}{$K$} & NuS & ADV-NuS \\ \cmidrule{2-3} \cmidrule{5-6} \cmidrule{8-9}
                        & \multicolumn{1}{c}{C.R.} & \multicolumn{1}{c}{C.R.} &                      & \multicolumn{1}{c}{C.R.} & \multicolumn{1}{c}{C.R.} &        & C.R. & C.R. \\
\midrule
0.1 & 0.08 & 0.81 & 10 & 0.08 & 0.86 & 100 & 0.07 & 0.73 \\
0.2 & 0.07 & 0.82 & 20 & 0.07 & 0.83 & 50 & 0.07 &  0.87 \\
0.3 & 0.09 & 0.90 & 30 & 0.08 & 0.82 & 25 & 0.09 & 1.01 \\
0.4 & 0.09 & 0.95 & 40 & 0.07 & 0.73 & 10 & 0.09 & 1.00 \\
0.5 & 0.10 & 1.12 & 50 & 0.07 & 0.73 & - & - & - \\
\bottomrule
\end{tabular}\label{last}
\end{table}

\subsubsection{Sensitivity of Hyper-parameters in GuideFlow}

\noindent \textbf{Impact of $\lambda$.} As the $\lambda$ increases from 0.1 to 0.5, the collision rates of GuideFlow on both NuScenes and ADV-NuScenes datasets demonstrate a upward trend. The performance degradation stems not from the constraint strategy itself, but from excessive interference with the predicted velocity field, which compromises the smoothness of the flow and reduces trajectory quality as shown in Tab.~\ref{last}.

\noindent \textbf{Impact of $k_c$.}
As shown in Tab.~\ref{last}, the observed continuous decline in collision rate as k increases from 10 to 50 confirms the effectiveness of the CF module. This trend indicates that applying constraints nearer to the endpoint of flow matching provides stronger and more direct guidance for trajectory generation, thereby resulting in a higher probability of satisfying the desired constraints.

\noindent \textbf{Impact of $K$.}
As the number of sampling steps decreases, GuideFlow exhibits a consistent decline in performance. This phenomenon can be attributed to the fact that reducing the number of sampling steps corresponds to increasing the step size. This strategy only preserves sampling fidelity when the probability flow is a ``straight" flow. However, since the learned probability flow in the GuideFlow is not ``straight", enlarging the step size directly disrupts the sampling process, leading to performance degradation, as shown in Tab.~\ref{last}.

\subsection{More Visualizations of Planning Results}
\label{sec:Qualitative}
To better illustrate the exceptional planning capabilities of GuideFlow, we present visualizations of its planning outcomes across multiple complex traffic scenarios, including turning maneuvers, congested traffic, and vehicle avoidance situations. We provide four qualitative results: (1) Planning under scenarios in NuScenes, (2) Planning under adversarial scenarios in ADV-NuScenes, (3) Closed-loop planning under scenarios in Bench2Drive and (4) Style transitions of planning guided by varying EP rewards. It is noteworthy that GuideFlow even \textcolor{red}{surpasses the expert(GT) trajectories} in certain scenarios of both the NuScenes and ADV-NuScenes datasets.

\begin{figure*}[t]
    \centering
    \includegraphics[width=1.0\linewidth]{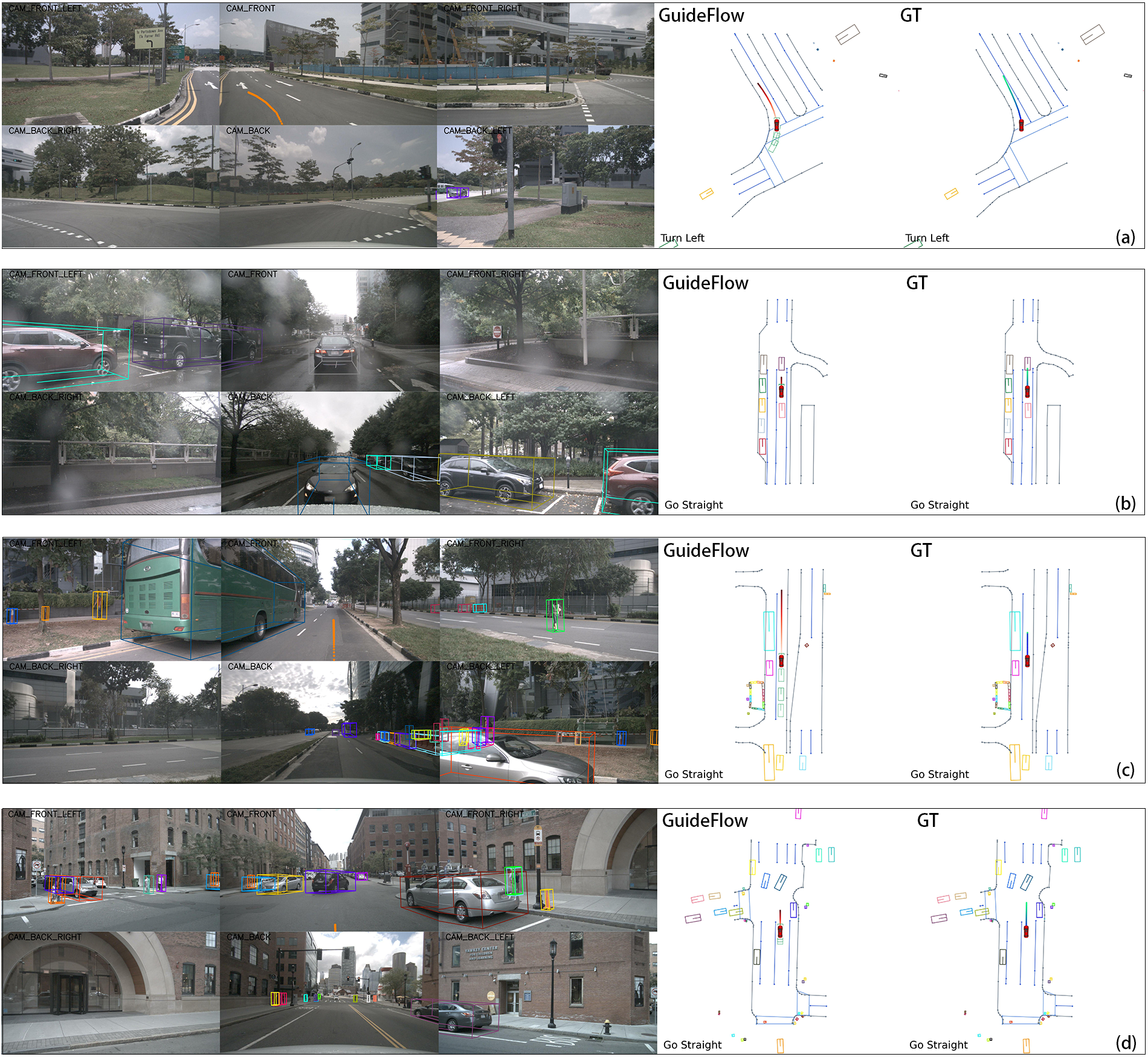}
    \caption{\textbf{\textcolor{red}{Better than expert trajectory.}} Visualization results in NuScenes dataset. (a) Turning scenario on a clear road. (b) Car following scenario, need maintain a safe distance from the leading vehicle to avoid collisions. (c) Straight road driving scenario, free of vehicles ahead. (d) Straight road driving scenario, a vehicle ahead is executing a turn, need the ego vehicle to give way.}
    \label{fig:nus}
\end{figure*}

\noindent \textbf{Planning under scenarios in NuScenes.}
As shown in Fig.~\ref{fig:nus}, we demonstrate the planning capability of GuideFlow across multiple scenarios from the NuScenes dataset. In subfigure (b), compared to the GT trajectory, GuideFlow adopts a more conservative car following strategy to mitigate the risk of colliding with the preceding vehicle. This behavioral discrepancy highlights the advantage of GuideFlow's non imitation learning training framework and constrained generation capability. A similar tendency is observed in subfigure (d). It is worth noting that in subfigure (c), where no vehicle is present ahead, GuideFlow exhibits more aggressive maneuvering. It further illustrates GuideFlow's capacity to dynamically adapt to varying scene.

\begin{figure*}[t]
    \centering
    \includegraphics[width=1.0\linewidth]{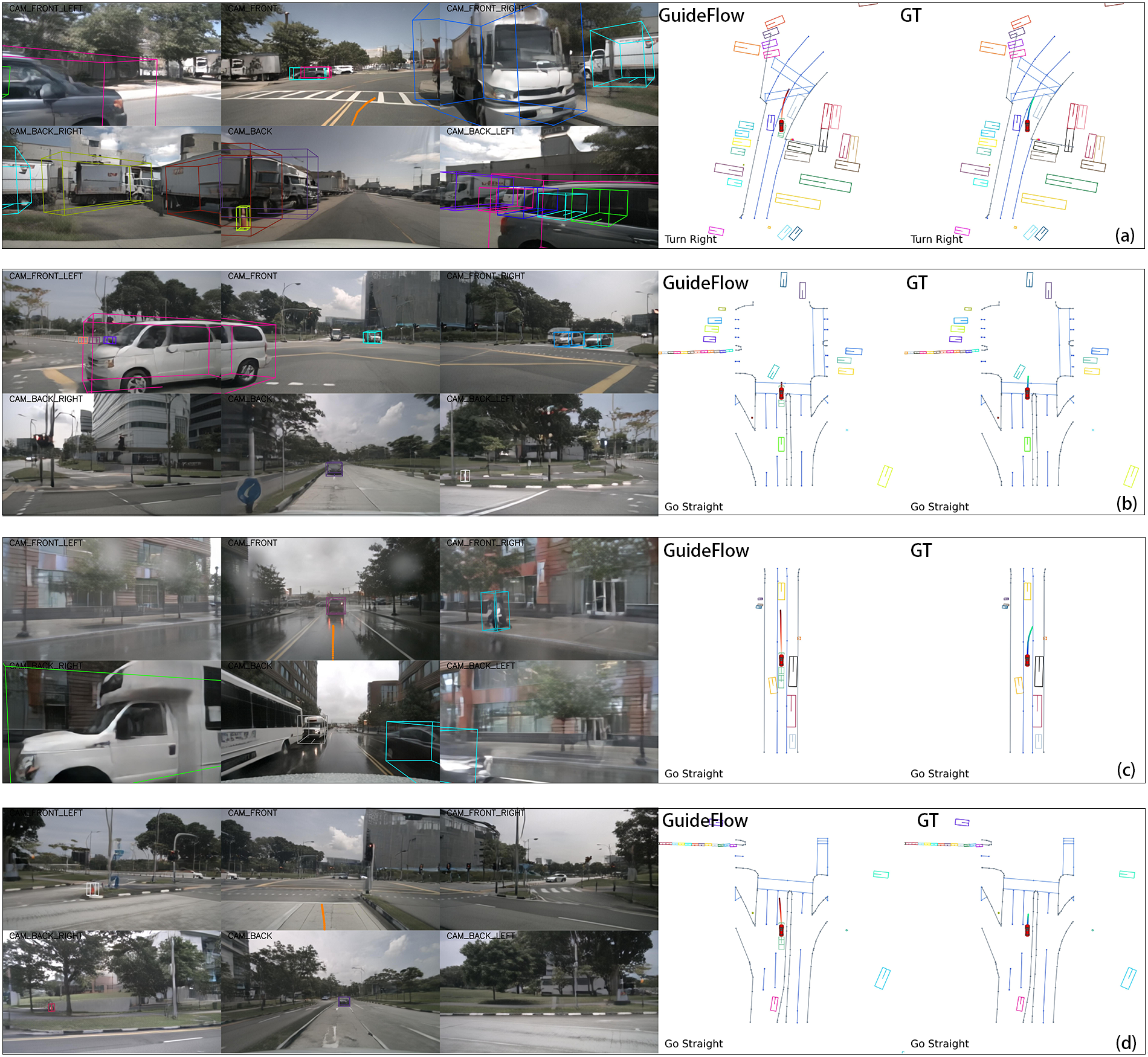}
    \caption{\textbf{\textcolor{red}{Better than expert trajectory.}} Visualization results in ADV-NuScenes dataset. (a) Emergency evasion scenario, a vehicle is encroaching into the lane, necessitating an immediate evasive maneuver by the ego vehicle. (b) Straight road driving scenario, a vehicle ahead is executing a turn, need the ego vehicle to give way. (c) Car following scenario, need maintain a safe distance from the leading vehicle to avoid collisions. (d) Straight road scenario, need ego vehicle avoid encroaching upon the non drivable area on its right.}
    \label{fig:adv-nus}
\end{figure*}

\noindent \textbf{Planning under adversarial scenarios in ADV-NuScenes.}
Fig.~\ref{fig:adv-nus} illustrates the performance of GuideFlow under various adversarial driving scenarios from the ADV-NuScenes dataset. In subfigure (a), when a vehicle illegally occupies the adjacent lane, GuideFlow demonstrates a timely avoidance strategy. In subfigure (b), compared to GT trajectory, GuideFlow exhibits a more proactive deceleration behavior to prevent a potential collision. Furthermore, in subfigure (d), to avoid encroaching into the non drivable area on the right, GuideFlow executes a pronounced evasion maneuver.

\begin{figure*}[t]
    \centering
    \includegraphics[width=1.0\linewidth]{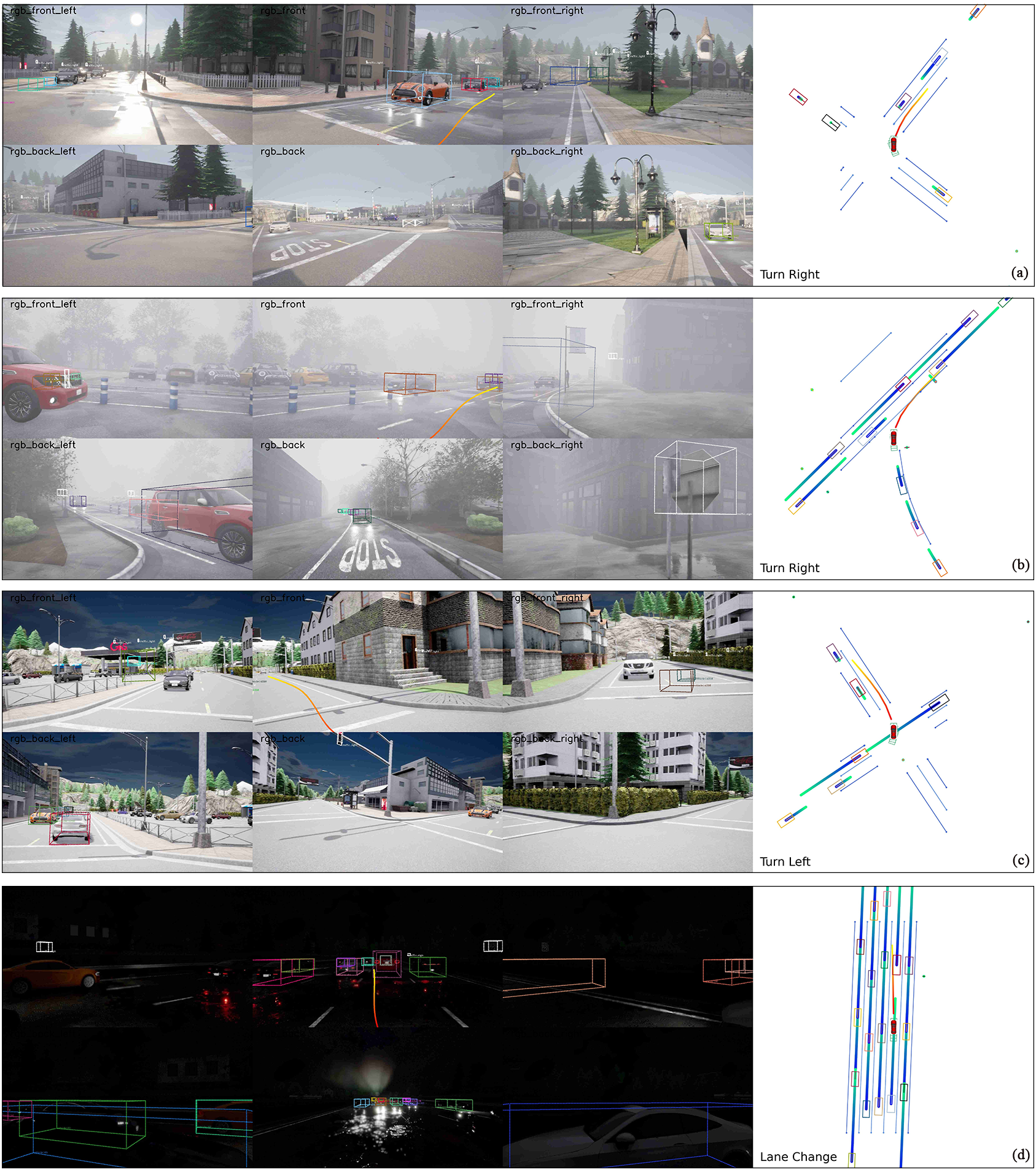}
    \caption{\textbf{Visualization results (Closed loop) in Bench2Drive dataset.} (a) Turning scenario on a clear road. (b) Turning scenario in congested traffic, need the ego vehicle avoid a collision. (c) Turning scenario in congested traffic, need the ego vehicle avoid a collision. (d) Lane change scenario in congested traffic, need the ego vehicle avoid a collision.}
    \label{fig:ben}
\end{figure*}

\noindent \textbf{Closed-loop planning under scenarios in Bench2Drive.}
To evaluate the closed loop planning capability of GuideFlow, we visualized its planning performance in the CARLA simulator as shown in Fig.~\ref{fig:ben}. The results demonstrate that GuideFlow consistently produces smooth and safe trajectories during various maneuvers, including turns and lane changes. Notably, it maintains reliable planning performance even under adverse visual conditions, such as foggy weather and nighttime.

\begin{figure*}[t]
    \centering
    \includegraphics[width=1.0\linewidth]{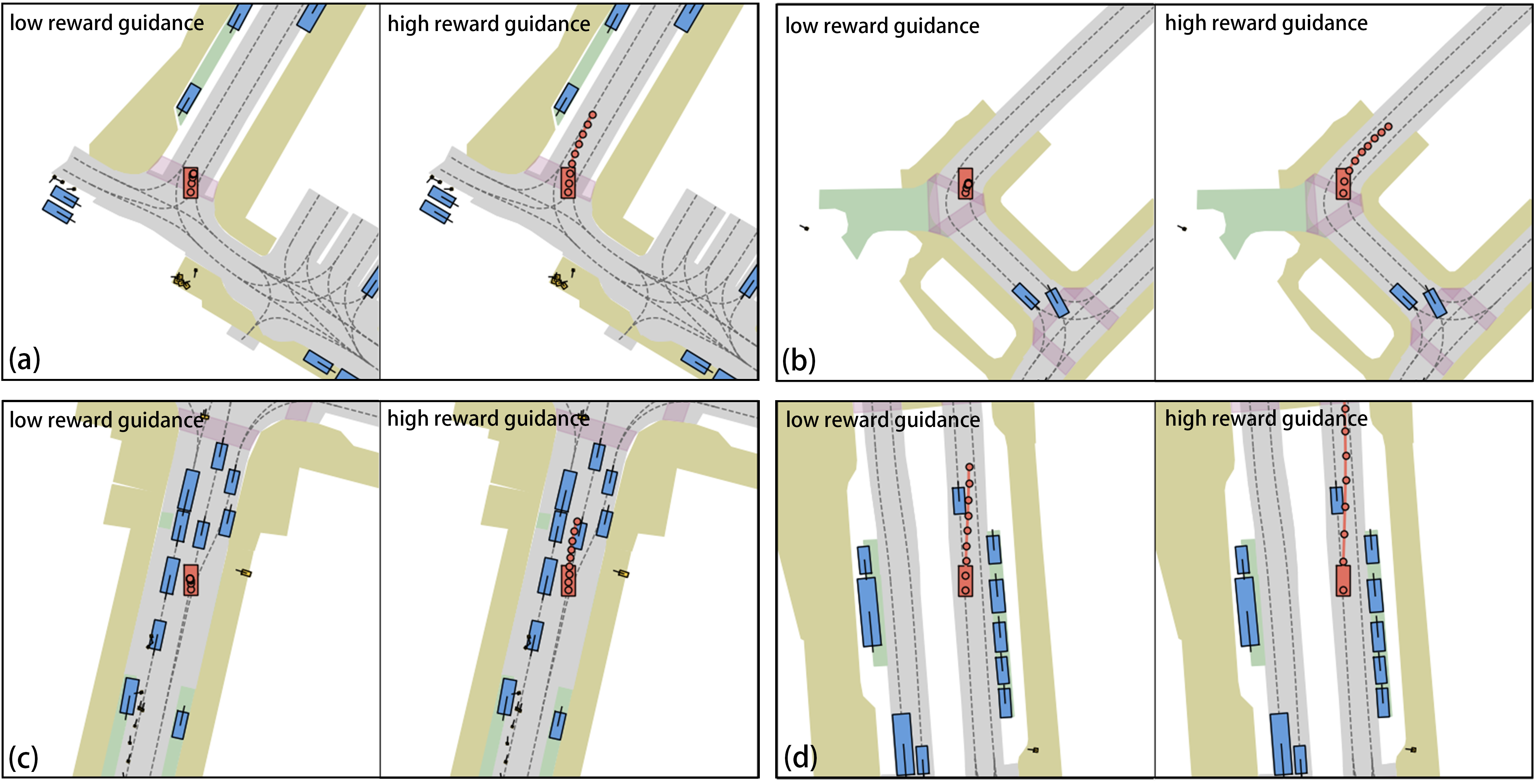}
    \caption{\textbf{Visualization of Trajectory Style Transitions on the NavSim Dataset.} The left side of each subfigure corresponds to trajectory generation guided by a low EP reward, while the right side corresponds to generation under a high EP reward.}
    \label{fig:ep}
\end{figure*}

\noindent \textbf{Style transitions of planning guided by varying EP rewards.}
Fig.~\ref{fig:ep} illustrates the trajectory style transitions capability of GuideFlow. When guided by a low EP reward value, the system generates trajectories with conservative driving behavior. Conversely, under high EP reward guidance, it exhibits more aggressive maneuvering, such as overtaking strategies, as exemplified in subfigures (c) and (d) of Fig.~\ref{fig:ep}.


{
    \small
    \bibliographystyle{ieeenat_fullname}
    \bibliography{main}
}
